\documentclass[a4paper,11pt]{style}

\usepackage{amsmath,amsfonts,amssymb,amsthm}
\usepackage{mathtools}
\usepackage{bbm}
\usepackage{accents}

\usepackage{comment}
\usepackage{adjustbox}
\usepackage{tikz}
\usepackage{overpic}
\usepackage{enumitem}
\usepackage{rotating}
\usepackage{algorithm}
\usepackage{algpseudocode}
\usepackage[textsize=tiny]{todonotes}
\usepackage[table]{xcolor}

\usepackage{amsmath,amsfonts,amssymb,amsthm}

\usepackage{amsmath}
\usepackage{amssymb}
\usepackage{mathtools}
\usepackage{amsthm}
\usepackage{natbib}
\usepackage{xcolor}

\usepackage[most]{tcolorbox}

\definecolor{sand}{RGB}{248,245,235}  
\definecolor{sandborder}{RGB}{220,215,200}
\definecolor{darkblue}{RGB}{40,70,140}   
\definecolor{parchment}{RGB}{252,249,240}
\definecolor{parchmentborder}{RGB}{222,215,195}
\definecolor{sage}{RGB}{244,248,244}
\definecolor{sageborder}{RGB}{200,210,200}
\definecolor{lavender}{RGB}{247,246,250}
\definecolor{lavenderborder}{RGB}{210,205,225}

\theoremstyle{plain}

\newtheorem{proposition}{Proposition}
\newtheorem{lemma}{Lemma}
\newtheorem{theorem}{Theorem}

\newtheorem{corollary}{Corollary}

\newcommand{\R}{\mathbb{R}}

\usepackage[textsize=tiny]{todonotes}

\title{Coarse-Graining Optimal Transport for  Rollout-Free Post-Training of Flow Matching}

\renewcommand{\authorlist}{%
  Soon Hoe Lim\textsuperscript{*}\textsuperscript{1,2}
  \hspace{0.35cm}
  Aditi Gupta\textsuperscript{4}
  \\[3pt]
  Michael W. Mahoney\textsuperscript{3,4,5}
  \hspace{0.35cm}
  N. Benjamin Erichson\textsuperscript{4,5}
}

\renewcommand{\affiliationlist}{%
  {\small
    \textsuperscript{1}Department of Mathematics,
    KTH Royal Institute of Technology
  }\\[1.5pt]
  {\small
    \textsuperscript{2}Nordita,
    KTH Royal Institute of Technology and Stockholm University
  }\\[1.5pt]
  {\small
    \textsuperscript{3}Department of Statistics,
    University of California at Berkeley
  }\\[1.5pt]
  {\small
  \textsuperscript{4}International Computer Science Institute
  }\\[1.5pt]
  {\small
    \textsuperscript{5}Lawrence Berkeley National Laboratory
  }%
}

\begin{document}

\abstract{%
Flow matching depends critically on the coupling used to pair source and target samples. 
Optimal transport (OT) couplings can reduce velocity conflict and straighten trajectories; but, in practice, challenges remain: the source comes from a continuous distribution, only a finite number of target samples data are accessible, and the neural network has finite capacity. 
For these reasons, making the coupling more OT-like may \emph{not} consistently improve sampling quality, in particular in the low number-of-function-evaluation (NFE) regime.
Motivated by this, we introduce ReCouple, a controllable framework that constructs coarse-grained OT couplings, to better understand rollout-free OT-based post-training. 
We also consider extensions of this basic framework:
ReCouple+ extends this toward the continuous-source setting by increasing source resolution, converging to semi-discrete OT in the appropriate limit; and 
ReCouple++ further adds spatial regularization, motivated by the observation that straight trajectories need not correspond to spatially simple velocity fields. 
Empirical validation demonstrates the effectiveness of these methods, and it reveals an NFE-dependent interplay between coupling optimality, neural realizability, and generation quality.\\

{\sffamily\bfseries\color{muxfg}Code: } {\color{blue}{\url{https://github.com/shoelim/recouple}}}
}

\maketitle

\begingroup
\renewcommand{\thefootnote}{*}
\footnotetext{Work done while visiting ICSI/UC Berkeley.}
\endgroup

\section{Introduction}
Flow matching (FM) \citep{lipman2022flow, liu2022flow, albergo2023stochastic} provides an elegant generative modeling framework based on dynamical transport between probability distributions. 
Given a prescribed probability path from a simple source distribution to the target distribution, FM learns the associated velocity field with a neural network and generates samples by integrating the resulting ordinary differential equation (ODE). 
However, accurately resolving the learned flow often needs a large number of function evaluations (NFEs), requiring many neural network evaluations, particularly so when the flow trajectories are ``curved.''

The ``curvature'' or ``straightness'' of a flow is determined by the coupling between the source and target, which we make precise here.
For probability measures $\nu, \mu \in \mathcal{P}_2(\R^d)$ (i.e., those with finite second moments), we  let $\Pi(\nu,\mu)$ denote the set of couplings with marginals $\nu$ and $\mu$.  
Given a coupling distribution $\gamma \in \Pi(\nu, \mu)$, consider the source-target pair $(Z,Y)\sim\gamma$, and the probability path $X_t = (1-t) Z + t Y$, $t \in [0,1]$. 
Conditioned on the pair $(Z,Y)$, the path is a straight line, and $\dot{X}_t = Y - Z$. 
The regression target in FM is the velocity field $v^*_t(x; \gamma) = \mathbb{E}_\gamma [\dot{X}_t | X_t = x]$. 
Thus, the target depends not only on the probability path, but also on the choice of the coupling.

A simple choice is a random coupling that pairs $Z$ and $Y$ independently.
Although each conditional path is straight, the marginal vector field averages velocities from different source-target pairs, often resulting in curved flows. To see this, the interpolation $X_t$ induces a distribution of endpoint velocities $U=X_1-X_0$ conditional on the intermediate state $X_t=x$, whose mean is the marginal velocity $v_t^*(x; \gamma)$. 
Averaging conflicting velocities locally in space produces a velocity field whose direction can vary along trajectories over time, resulting in curved paths and larger discretization error at low NFEs \citep{liu2022flow}. When the conditional velocity distribution is concentrated near a single velocity, this local ambiguity is reduced and the induced flow is easier to integrate.

\begin{figure}[!t]
    \centering
        \caption{Generated samples on the 2D multimodal task across methods and NFEs.  Post-training using the ReCouple family improves recovery of the target modes, particularly at low NFEs.}
        \vspace{0.15cm}
\includegraphics[width=1\textwidth]{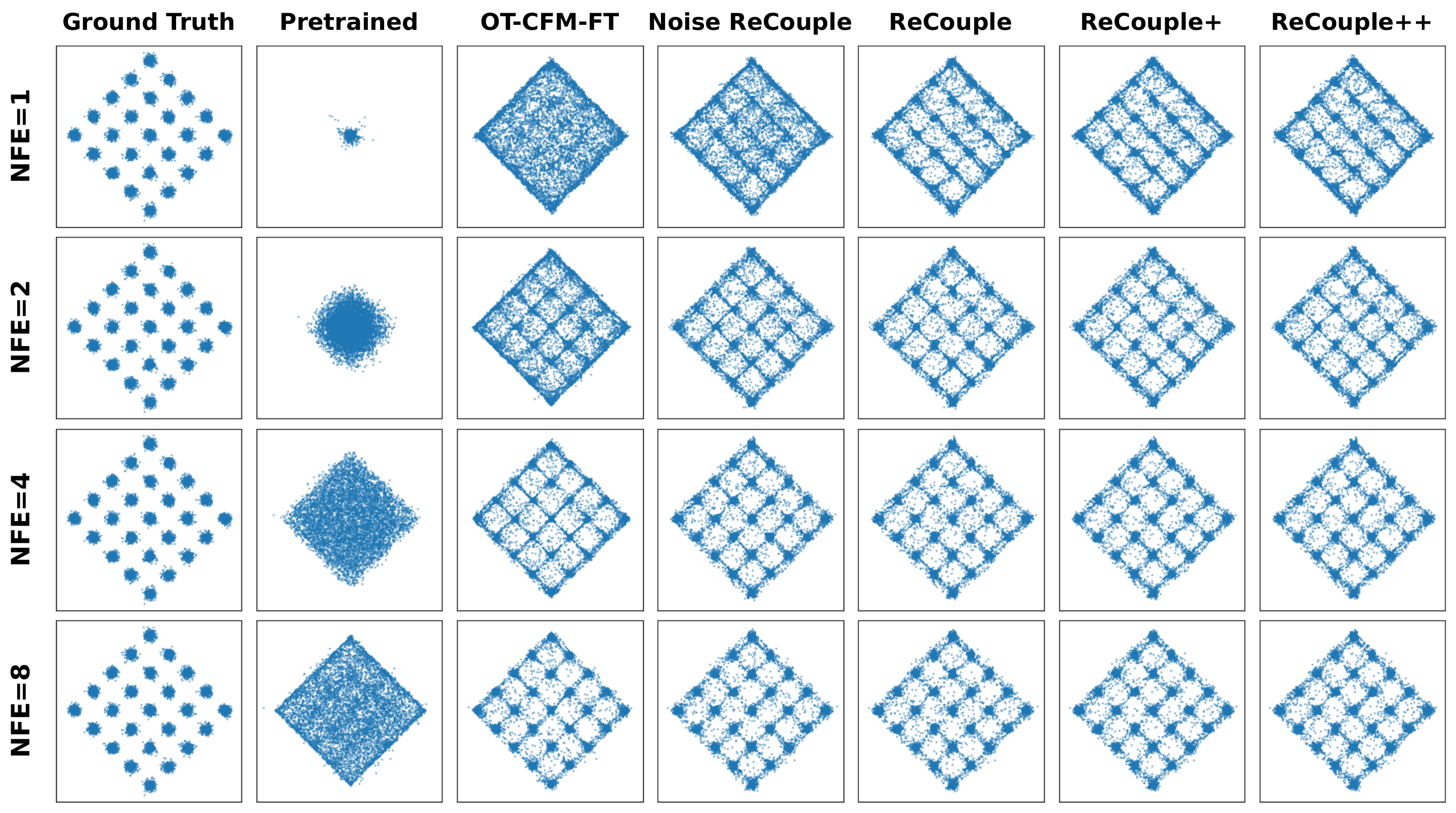}
    \label{fig:main_visual}
    \vspace{-0.5cm}
\end{figure}

Optimal transport (OT) instead chooses the minimum-cost pairing, giving the OT plan
$\gamma_{\mathrm{OT}}\in\mathrm{argmin}_{\gamma\in\Pi(\nu,\mu)}\mathbb{E}_{\gamma}[\|Z-Y\|^2]$ for quadratic cost (also  equivalent to maximizing endpoint alignment). 
Under quadratic OT, the induced displacement interpolation is a Wasserstein geodesic, so particles move with constant velocity along straight characteristics. 
Global OT over a large dataset is computationally costly, with standard  solvers scaling cubically in sample size.
In practice, batchwise OT solves OT only within each minibatch at each training iteration (\citet{tong2023improving, pooladian2023multisample}). 
However, this is suboptimal: each minibatch only covers a small region of the full distribution, so the batch-level couplings remain unstable across iterations and may result in curved flows. 
Due to these practical issues, OT-guided models are not widely used, and state-of-the-art models (e.g., \citet{liu2022flow, esser2024scaling}) are typically trained on independent coupling pairs, often resulting in curved flows and requiring many NFEs to resolve the curvature.

OT can lead to straight flows realized by the Brenier map \citep{kornilov2024optimal}, but in practice we work with finite number of training/target samples whereas the source distribution is absolutely continuous (e.g., Gaussian), thus solving an OT problem which is semi-discrete in nature. 
However, both OT and semi-discrete OT (SDOT) are  computationally costly \citep{peyre2025optimal}, while the  Brenier map can be hard for a finite-capacity neural network to realize \citep{pierret2026tessellations}. 
Moreover, even if we can approximate OT accurately, training naively on the OT pairs may not lead to a good few-step generator across all NFE regimes \citep{hui2025not}. 

Despite these challenges, OT remains attractive  for training and fine-tuning  FM models. 
However, what are the caveats, and how far can this push a given pretrained FM model toward an improved few-step generator? 
This in turn exposes a basic question: 
\begin{tcolorbox}[
  colback=parchment,
  colframe=sandborder,
  boxrule=0.3pt,
  left=8pt,
  right=8pt,
  top=1pt,
  bottom=1pt
]
\centering
What makes a source-target coupling good for flow matching training in practice? 
\end{tcolorbox}

In this work, we introduce a controllable OT approximation framework to systematically understand the tradeoffs involved. 
To better expose the challenge and isolate approximation effects, we focus on post-training, and we default to a single post-training scheme (see Algorithm \ref{alg:main_posttrain-coupling}), avoiding adding various training ``tricks'' \citep{esser2024scaling}.
These tricks can improve generative performance at the expense of higher computational cost, but they are often heuristic, task- and architecture-dependent (and thus less amenable to theory). 
Our scheme also does not require pretrained-model rollouts or teacher trajectories, unlike Reflow \citep{liu2022flow, kim2025simple}  and distillation-based schemes. It is also model-independent during coupling construction, unlike model-aligned approaches such as MAC \citep{lin2026beyond} that use the current model to assess candidate pairs; see App.~\ref{app_relatedwork} for a discussion of related work.   
Throughout, we use explicit Euler as the default sampler at inference, so that NFE equals the number of sampling steps. Our focus is on improvements arising from the training procedure rather than from the choice of numerical solver.

Our main contributions are as follows.
\begin{itemize}[leftmargin=10pt]
    \item 
    {\bf ReCouple: an approximate OT scheme.} 
    We introduce and study ReCouple, a divide-and-conquer approximation to OT based on structured blockwise assignment. This induces a natural {\it coarse-graining} of the transport problem: the block size $B$ controls its resolution, ranging from projection-based rank matching at $B=1$ to global empirical OT at $B=N$.
    We further provide mechanistic insights to understand ReCouple and its core components; see  App.~\ref{app_recouple}.
    
    \item 
    {\bf ReCouple+: increasing the source resolution.} 
     To increase source resolution, ReCouple+ replaces the one-to-one block assignments in ReCouple with pooled capacity-$M$ assignments using $MN$ source samples for $N$ targets, separating geometric locality from continuous-source approximation. 
    ReCouple+ thus serves as a controlled bridge between discrete OT and semi-discrete OT (SDOT), 
    and we further show its convergence to SDOT in the appropriate limit (see App.~\ref{app_proof}).
        
    \item {\bf ReCouple++: introducing spatial regularization.} 
    We observe empirically that approaching OT more closely needs not make the velocity field easier to learn. 
   In particular, straighter transport trajectories can still correspond to spatially sharp Eulerian velocity fields with large Jacobians. Motivated by this distinction between trajectory straightness and spatial representational complexity, ReCouple++ augments ReCouple+ with spatial Jacobian regularization.  
   
    \item {\bf Empirical validations.} Empirical validations on multimodal 2D data and  image generation tasks demonstrate the effectiveness of the ReCouple family compared to similar baselines. It also exposes the lesson that transport optimality and neural realizability need not improve sample quality together consistently across all NFE regimes. ReCouple++ turns the insights into an improved post-trained model by jointly controlling OT plan's approximation and spatial field complexity.
\end{itemize}

\section{Flow Matching and Optimal Transport}
\subsection{Conditional Flow Matching (CFM)}
Given a source-target pair $(Z,Y)\sim\gamma\in\Pi(\nu,\mu)$, define $X_t = (1-t)Z+tY$ and $U=Y-Z$.

\noindent {\bf CFM Training objective.} 
The population CFM loss \citep{tong2023improving} is:
\begin{equation}
\label{eq:posttrain}
\mathcal L_{\rm CFM}
(w;\gamma)
=
\mathbb E_{\substack{
(Z,Y)\sim \gamma\\
t\sim \mathrm{Unif}[0,1]
}}
\left[
\left\|
v^w_t((1-t)Z+tY)
-(Y-Z)
\right\|^2
\right].
\end{equation}
We can decompose the loss into 
$\mathcal L_{\rm CFM}(w;\gamma) =  \mathcal{E}(w; \gamma) + \mathcal{A}(\gamma),$
where 
$$\mathcal{E}(w; \gamma) := \mathbb E_{
(Z,Y)\sim \gamma, \ 
t\sim \mathrm{Unif}[0,1]} \|v_t^w(X_t) - v^*_t(X_t; \gamma) \|^2$$
quantifies the field realization error and 
$$\mathcal{A}(\gamma) := \mathbb E_{
(Z,Y)\sim \gamma, \ 
t\sim \mathrm{Unif}[0,1]}[\operatorname{tr} \Sigma_t^\gamma(X_t)]$$  
quantifies the irreducible error. 
Here, $v_t^*(x; \gamma) :=
\mathbb E_\gamma[\dot X_t\mid X_t=x]$, the conditional covariance $\Sigma_t^\gamma(x) := \operatorname{Cov}_\gamma(\dot{X}_t \mid X_t=x)$ quantifies the disagreement among pathwise velocities passing through $(t,x)$, with $\operatorname{tr}\Sigma_t^\gamma(x)=\mathbb E_\gamma[ \|\dot{X}_t -v^*_t(x; \gamma)\|^2\mid X_t=x]$. 
Thus, large $\Sigma_t^\gamma$ corresponds to conflicting regression targets, and $\mathcal{A}(\gamma)$ reflects the ambiguity induced by the coupling. 

\subsection{Semi-Discrete OT}
 
\paragraph{Semi-discrete optimal transport (SDOT)} 
Given an absolutely continuous (with respect to Lebesgue measure) source distribution $\nu \in \mathcal{P}_2(\R^d)$ and a discrete empirical data distribution $\hat{\mu}_N = \frac{1}{N} \sum_{i=1}^N \delta_{y_i}$, where $N$ is the sample size and $y_i \overset{\text{iid}}{\sim} \mu$ for some unknown target distribution, SDOT finds the optimal coupling \citep{peyre2025optimal}:
\begin{equation}
\gamma_N^\star
\in
\mathrm{argmin}_{\gamma\in\Pi(\nu,\widehat\mu_N)}
\int\|z-y\|^2\, \mathrm{d} \gamma(z,y).
\label{eq:empirical-sdot}
\end{equation}

\paragraph{Why SDOT is desirable}
In practice, we are given access to only a finite number of training samples and we must learn the FM model using them. 
Thus, the CFM training operates in the data setting of SDOT, where the source distribution is continuous while the target distribution is discrete. 

By Brenier's theorem \citep{villani2021topics}, there exists a 
$\nu$-a.e. unique OT map via a potential function~$\phi$:
\begin{equation}
    T_N^\star=\nabla\phi,
    \qquad
    (T_N^\star)_\#\nu=\widehat\mu_N,
    \qquad
    \gamma_N^\star=(\mathrm{Id},T_N^\star)_\#\nu .
\label{eq:sdot-map}
\end{equation}
Since $\widehat\mu_N$ is discrete,
$T_N^\star$ is piecewise constant. 
In particular, there exist weights
$\psi=(\psi_1,\ldots,\psi_N)$ such that the Laguerre cells (see Section 6.3 in \citep{peyre2025optimal})
\begin{equation}
    L_i(\psi)
    =
    \left\{
        z:
        \frac12\|z-y_i\|^2-\psi_i
        \leq
        \frac12\|z-y_j\|^2-\psi_j,
        \ \forall j
    \right\}
\label{eq:laguerre-cells}
\end{equation}
satisfy $\nu(L_i)=1/N$, and
$T_N^\star(z)=y_i$ for $\nu$-a.e. $z\in L_i$. 
The map $T_N^*$ is discontinuous at the boundaries of each cell.

The following proposition highlights two favorable properties of the SDOT flow. Its deterministic coupling eliminates conditional velocity conflict, while its material acceleration,
$D_t v_t:=\partial_t v_t+(v_t\cdot\nabla)v_t$, vanishes along the displacement interpolation. Consequently, the velocity is constant along each characteristic, and a single Euler step from $t=0$ to $t=1$ recovers the SDOT map exactly.

\begin{proposition}[Straightness of the SDOT flow]
\label{prop:sdot-straight}
Let $T=T_N^\star$ be the Brenier map transporting $\nu$ to
$\widehat\mu_N$, and define the affine interpolation
$X_t=(1-t)Z+tT(Z)$, $\dot{X}_t=T(Z)-Z$, where $Z\sim\nu.$
Let
$v_t^*(x)
    :=
    \mathbb E[\dot X_t\mid X_t=x]$
denote the corresponding marginal velocity field. Then, for every
$0\leq t<1$,
$\operatorname{Cov}(\dot X_t\mid X_t)=0$,
and hence $v_t^*(X_t) =
    \dot X_t
    =
    T(Z)-Z 
     \ \text{a.s..}$
In particular, the velocity is constant along each characteristic,
\begin{equation}
    \frac{d}{dt}v_t^*(X_t)= \partial_t v_t^*
    +
    (v_t^*\cdot\nabla_x)v_t^* = 0,
\end{equation}
wherever the derivatives exist. 
\end{proposition}

Hence,  the SDOT map provides a principled target for few-step generation in FM. 
Degradation at low NFEs thus arises from approximating the OT velocity field over $t \in [0,1]$, rather than from intrinsic velocity conflict or curvature of the underlying flow trajectories. Approximate OT constructions (such as ReCouple and
ReCouple+; see later subsections) can therefore be viewed as controlled approximations to this target map, allowing us to study how much of the OT geometry is needed at different NFE budgets. 
On the other hand, learning the discontinuous SDOT map with a spatially smooth velocity field can be difficult, thus at odds with the goal of exact one-step integration  (see ReCouple++ in a later subsection).

\section{ReCouple}

\begin{algorithm}[!t]
\caption{ReCouple}
\label{alg:recouple-main}
\begin{algorithmic}[1]
\Require target bank $y_{1:N}$; source bank $z_{1:N}$; block size $B \leq N$ ($B$ divides $N$); proposal count $R$
\State Add the identity pairing $\pi_0(i)=i$ to the candidate set.
\For{$r=1,\ldots,R$}
    \State Draw $\theta_r\sim\mathrm{Unif}(\mathbb S^{d-1})$.
    \State Let $\sigma_r,\tau_r\in S_N$ sort
    $\{\langle\theta_r,y_i\rangle\}_{i=1}^N$ and
    $\{\langle\theta_r,z_j\rangle\}_{j=1}^N$, respectively 
    \State Partition the ordered positions into
    $I_b=\{(b-1)B+1,\ldots,bB\}$, $b=1,\ldots,N/B$.
    \For{$b=1,\ldots,N/B$}
        \State Compute
        \[
        \rho_{r,b}\in\mathrm{argmin}_{\rho\in\mathfrak S(I_b)}
        \sum_{k\in I_b}
        \left\|
        y_{\sigma_r(k)}
        -
        z_{\tau_r(\rho(k))}
        \right\|^2.
        \]
    \EndFor
    \State Define $\pi_r\in  S_N$ by
    $\pi_r(\sigma_r(k))=\tau_r(\rho_{r,b}(k))$ for $k\in I_b$.
    \State Evaluate
    $C_N(\pi_r)=\frac{1}{N}\sum_{i=1}^N
    \|y_i-z_{\pi_r(i)}\|^2$.
\EndFor
\State Select
$r^\star\in\arg\min_{r=0,\ldots,R} C_N(\pi_r)$
and set $\widehat\pi=\pi_{r^\star}$.
\State \Return
$\widehat\gamma
=
\frac{1}{N}\sum_{i=1}^N
\delta_{(z_{\widehat\pi(i)},y_i)}$.
\end{algorithmic}
\end{algorithm}

\subsection{ReCouple}

{\bf How much OT is enough? } 
We introduce  a controllable framework to better understand this. The key to this framework is ReCouple (see  Algo.~\ref{alg:recouple-main} and Algo.~\ref{alg:recouple-general} for details), an approximate OT scheme that, at a high level, works as follows.
We repeatedly draw a random direction, sort both source and target banks by projection onto the random direction, split the ordered lists into corresponding blocks of size $B$, solve the linear sum assignment (LSA) problem inside each block, and add the concatenated permutation into the candidate set. The final permutation is chosen  as the one that realizes the minimal cost among the candidates.
Each block-local LSA can be solved via the Hungarian or the modified Jonker–Volgenant algorithm \citep{crouse2016implementing}. Importantly:

\begin{tcolorbox}[
  colback=parchment,
  colframe=sandborder,
  boxrule=0.1pt,
  left=8pt,
  right=8pt,
  top=0.4pt,
  bottom=0.4pt
]
\centering
$B$ controls the ``coarse-graining scale" of the  ordering induced by a 1D projection: after sorting source and target samples by their projected coordinates, consecutive groups of $B$ samples define corresponding blocks within which the full OT is solved.
\end{tcolorbox}
See App.~\ref{app_recouple} for additional insights. Throughout, $S_N$ denotes the set of permutations of $\{1,\ldots,N\}$.

\paragraph{ReCouple vs. SWGG and global empirical OT} 
ReCouple's plan can be viewed as a blockwise relaxation of the plan computed by minimizing SWGG \citep{mahey2023fast}, with $B=1$ recovering the SWGG's plan associated with the selected direction (see App.~\ref{app_recouple} for details), while $B=N$ recovers the global empirical OT coupling. 
By solving only local assignment problems induced by 1D random projections, ReCouple interpolates between the computational efficiency of SWGG-based plan computation and the geometric fidelity of global OT, reducing the per-proposal assignment cost from $O(N^3)$ to $O(NB^2)$. 
For nested blockings obtained by merging aligned finer blocks, increasing $B$ enlarges the admissible set and cannot increase the blockwise OT cost.

\subsection{ReCouple+}
ReCouple+ is an enhanced version of ReCouple that increases the source resolution. In particular, the source bank is enlarged by a factor of $M$, in which case the LSA assignment problem extends to a capacity-$M$ assignment problem (see Algo.~\ref{alg:recouple-plus} for details).
Intuitively, increasing $M$ reduces the finite-source approximation error.

This approximation scheme is not designed to be the most efficient solver for large-scale OT \citep{genevay2016stochastic,genevay2018learning}. Rather, we exploit its explicit resolution parameters: $B$ controls the scope of the admissible transport, while $M$ controls the resolution of the continuous source approximation. Indeed, as $B$ and $M$ are increased toward the SDOT limit, the
computational advantage of the blockwise approximation diminishes.
Its purpose here is instead to provide an explicit and controllable
family of couplings through which these different transport resolutions
can be studied.

\paragraph{When does ReCouple+ approach SDOT}
The parameters $B$ and $M$ control two distinct approximations to the
SDOT problem. The block size $B$ controls the
\emph{geometric resolution} of the coupling by restricting transport to
corresponding projected-rank blocks, while the pooling factor $M$
controls the \emph{source resolution} by increasing the number of
samples used to approximate the continuous source distribution.
This distinction becomes clear at the two limits.  When $B=N$,
the block restriction disappears, and ReCouple+ solves the discrete OT
problem between the pooled empirical source
\[
\widehat\nu_{MN}
=
\frac{1}{MN}\sum_{j=1}^{MN}\delta_{Z_j}
\]
and the empirical target $\widehat\mu_N$.  As $M\to\infty$,
$\widehat\nu_{MN}$ converges to $\nu$ in $W_2$ almost surely, and the
ReCouple+ cost and optimal coupling converge to those of the SDOT
problem between $\nu$ and $\widehat\mu_N$.  In contrast, if $B<N$ is
fixed and $M\to\infty$, the limit remains a block-restricted SDOT
problem: increasing $M$ resolves the continuous source more accurately
but does not remove the geometric restriction imposed by $B$.
At the limiting corner $B=N$ and $M\to\infty$, both approximations are
removed and SDOT is recovered. See the theorem below, and App.~\ref{app_proof} for more results and  details.

\begin{theorem}[Informal: Convergence of ReCouple+ to SDOT]
\label{thm:fixed-N-SDOT}
Let $\widehat{\mu}_N
=
\frac{1}{N}\sum_{i=1}^N \delta_{y_i}$ be given and suppose that
$\nu\in\mathcal{P}_2(\mathbb{R}^d)$ is absolutely continuous.
Let $Z_i  \overset{\mathrm{iid}}{\sim}\nu,$
and, for each $M\ge1$, define
$\widehat{\nu}_{MN}
=
\frac{1}{MN}
\sum_{j=1}^{MN}\delta_{Z_j}.$
Consider ReCouple+ with one global block, $B=N$, and pooling factor
$M$. Let $\widehat{\gamma}_{N,M}^{+}$ denote the resulting empirical
coupling. 
Then, 
$$W_2(\widehat{\gamma}_{N,M}^{+}, \gamma_N^{\mathrm{SDOT}}) \to 0$$ 
almost surely,
where $\gamma_N^{\mathrm{SDOT}}$ is the unique optimal
coupling between $\nu$ and $\widehat{\mu}_N$.
\end{theorem}

\subsection{ReCouple++}
\label{subsec_recouple++}

\paragraph{Why ReCouple or ReCouple+ alone may  not be optimal}
While training on the ReCouple  pairs can make the learned flow straighter across $t \in [0,1]$, it does not explicitly address the spatial variation of the velocity field (either in data space or latent space) during training. 
In particular, it does not account for nearby trajectories with substantially different velocities, which may be difficult for a finite-capacity model to distinguish. 
To account for this finite spatial resolution, we introduce a scale $\sigma>0$ and define $W_t := X_t + \sigma \xi$, $\xi\sim\mathcal N(0,I)$, where $\xi$ is independent of $(Z,Y)$. For fixed $t$, define the time-local irreducibility error
$ \mathcal A_t(\gamma) := \mathbb E_{(Z,Y)\sim\gamma}
    \left[
        \operatorname{tr}\Sigma_t^\gamma(X_t)
    \right]$, and the Bayes risk:
\begin{equation}
\label{eq:resolution-risk}
    \mathcal D_{\gamma,\sigma}(t)
    :=
    \inf_f
    \mathbb E
    \left[
        \left\|
        \dot X_t-f(W_t)
        \right\|^2
    \right]
    =
    \mathbb E
    \left[
        \operatorname{tr}
        \operatorname{Cov}
        (\dot X_t\mid W_t)
    \right].
\end{equation}
Unlike $\mathcal A_t(\gamma)$, which only measures disagreement among
velocities that meet at the same state, $\mathcal D_{\gamma,\sigma}(t)$
also regards sufficiently nearby trajectories as locally indistinguishable.

Let $J_\gamma(t)^2 :=
        \mathbb E
        \left[
            \left\|
            \nabla_xv_t^*(X_t;\gamma)
            \right\|_F^2
        \right]
    $ for $t \in (0,1)$. 
    Assuming that, for fixed $t$, $X_t$ has a smooth positive density and
$v_t^*(\cdot;\gamma)$ is sufficiently smooth with suitable integrability
conditions, we have, as $\sigma\to 0$,
\begin{equation}
\label{eq:small-sigma}
    \mathcal D_{\gamma,\sigma}(t)
    =
    \mathcal A_t(\gamma)
    +
    \sigma^2 J_\gamma(t)^2
    +
    \mathcal O(\sigma^4).
\end{equation}

This suggests that suppressing the coupling ambiguity alone may not be enough, and adding spatial Jacobian regularization can be beneficial to improve ReCouple-based post-training. Therefore, we consider an enhanced version of ReCouple+, which we refer to as  ReCouple++ (we refer to a similarly enhanced version of ReCouple as ReCouple-AS), whose training is done by minimizing 
\begin{equation}
\mathcal{L}(w; \gamma) = \mathcal{L}_{CFM}(w; \gamma) + \lambda \  \mathbb{E}_{t \in \mathrm{Unif}[0,1]} \mathcal{R}_\sigma(w; t),    
\end{equation}
where 
\begin{equation}
\label{eq:spatial-reg}
    \mathcal R_\sigma(w;t)
    :=
    \frac{1}{2\sigma^2}
    \mathbb E
    \left[
        \left\|
        v_t^w(X_t+\sigma\xi)
        -
        v_t^w(X_t)
        \right\|^2
    \right].
\end{equation}
Assuming that $v_t^w$ is sufficiently smooth in $x$ with suitable
integrability conditions, a Taylor expansion yields, as $\sigma\to0$,
\begin{equation}
\label{eq:spatial-reg-limit}
    \mathcal R_\sigma(w;t)
    =
    \frac{1}{2}
    \mathbb{E}
    \left[
        \left\|
        \nabla_xv_t^w(X_t)
        \right\|_F^2
    \right]
    +
    \mathcal O(\sigma^2).
\end{equation}
We further anneal the penalty to zero so that the final training phase optimizes the unregularized CFM objective (see Alg. \ref{alg:posttrain-coupling} in App.~\ref{app_detailedalgo} for full details).

This issue becomes particularly relevant in the SDOT limit approached by ReCouple+ when $B=N$ and $M\to\infty$. Although SDOT displacement trajectories are exactly straight, the associated Eulerian velocity field is spatially sharp: within each transported Laguerre cell,
\begin{equation}
    \nabla_x v_t^\star=-\frac{1}{1-t}I_d,
\end{equation}
while globally stitching distinct cellwise branches requires Lipschitz scale at least $1/t$. Thus the SDOT velocity cannot remain uniformly spatially regular over time, motivating ReCouple++ to control the spatial variation of the learned field. See Theorem \ref{prop:app_straight-not-spatially-simple} in App.~\ref{app_proof}.

Meanwhile, by Proposition \ref{prop:sdot-straight},
$\frac{d}{dt}v_t^*(X_t)
= 0$.
In other words, the ideal SDOT has perfectly straight characteristics, but the  vector field is spatially sharp: it must distinguish many cells and needs to resolve increasingly steep transitions near the endpoints.
By penalizing spatially high-variation modes, the annealed regularizer biases the learned velocity field toward smoother realizations during the initial phase of post-training. As the penalty is annealed to zero, the final phase optimizes the original CFM objective without an explicit spatial penalty.

Notably, \citep{hui2025not}  studied this phenomenon in 3D point cloud generation and 
mitigate it by injecting noise into the source samples. We instead develop an explicit block/pooling hierarchy and regularize the spatial variation of the neural field without modifying the underlying pair bank. We therefore also compare against an analogous noise-injection variant (Noise-ReCouple) which perturbs the ReCouple source endpoint as
$z_\beta=\sqrt{1-\beta}\,z+\sqrt{\beta}\,\epsilon$,
with $\epsilon\sim\mathcal N(0,I)$, while keeping the target fixed.

\section{Empirical Validation}

In this section, we evaluate post-training of FM models on the ReCouple family and related baselines on 2D multimodal generation and image generation tasks (CIFAR-10 and ImageNet-256). Details, ablation studies, and additional results are provided in App.~\ref{app_experimental}.

\subsection{2D Multimodal  Generation}

We consider unconditional generation for a 2D multimodal dataset (see App.~\ref{app_2ddata} for details). The source samples are from the standard Gaussian, while the target samples are from a 25-mode Gaussian mixture. Fig. \ref{fig:main_visual} shows the samples generated by various baselines and the ReCouple schemes.

\begin{table}[!b]
\centering
\caption{Evaluation on the 2D multimodal task.
We report the 2-Wasserstein distance ($W_2$, lower is better) at different
NFEs, together with the neural
acceleration energy $\mathcal{E}_{\mathrm{acc}}$ (lower is better).
}
\label{tab:2d-main}
\vspace{0.1cm}
\small
\setlength{\tabcolsep}{4.5pt}
\begin{tabular}{lccccc}
\toprule
Method
& \multicolumn{4}{c}{$W_2 \downarrow$}
& $\mathcal{E}_{\mathrm{acc}} \downarrow$ \\
\cmidrule(lr){2-5}
& NFE 1 & NFE 2 & NFE 4 & NFE 8 & \\
\midrule
Pretrained
& 0.6103 & 0.4092 & 0.2119 & 0.1362 & 57.6858 \\

PersistentIndependent-FT
& 0.5924 & 0.4145 & 0.2164 & 0.1378 & 65.4474 \\

AlignFlow \citep{kong2026alignflow}
& 0.5234 & 0.3537 & 0.1880 & 0.1211 & 54.6471 \\

OT-CFM-FT \citep{tong2023improving}
& 0.0996 & 0.0875 & 0.0848 & 0.0859 & 1.3101 \\

SDOT \citep{mousavi2026flow}
& 0.1082 & 0.1092 & 0.1087 & 0.1087 & \textbf{0.2628} \\

Noise-ReCouple (c.f. \citep{hui2025not})
& 0.0895 & 0.0843 & 0.0849 & 0.0860 & 0.5316 \\

\rowcolor{orange!12} ReCouple
& 0.0866 & 0.0853 & 0.0857 & 0.0862 & 0.4461 \\

\rowcolor{orange!12} ReCouple-AS
& 0.0870 & 0.0848 & 0.0852 & 0.0858 & 0.3542 \\

\rowcolor{orange!12} ReCouple+
& \textbf{0.0859} & 0.0845 & 0.0848 & 0.0853 & 0.4401 \\

\rowcolor{orange!12} ReCouple++
& 0.0863 & \textbf{0.0840} & \textbf{0.0843} & \textbf{0.0849}
& 0.3642 \\
\bottomrule
\end{tabular}
\end{table}

From Table \ref{tab:2d-main}, we see that ReCouple substantially improves over the pretrained and independent-FT baselines, while ReCouple+ achieves the best $W_2$ at NFE~1 and ReCouple++ performs best at NFE 2-8. The improvements are also reflected in the other distributional metrics: the ReCouple variants achieve substantially lower sW2 than OT-CFM and SDOT, while maintaining high recall and coverage (see Table \ref{tab:2d-metrics-nfe1}-\ref{tab:2d-metrics-nfe8} in App.~\ref{app_2ddata}).

Spatial regularization provides a complementary control on realizability. It leads to consistently lower acceleration energy, while giving modest improvements at NFE 2-8 and little or no gain at NFE~1. Combining increased source resolution and spatial regularization in ReCouple++ gives the strongest overall $W_2$ performance beyond one step. Together, these results suggest that low-NFE quality depends on the interaction between the coupling,  neural realizability of the induced velocity field, and numerical discretization, rather than on OT optimality alone.

\subsection{CIFAR-10 Image Generation}

\begin{table}[!t]
\centering
\caption{CIFAR-10 FID results with the Rectified Flow pretrained
backbone. Lower is better. We fix $R=32$ and $B=2048$ for ReCouple,
ReCouple-AS, ReCouple+, and ReCouple++. We use $M=10$ for
ReCouple+ and ReCouple++. Results are averaged over three seeds.}
\label{tab:cifar-current-baselines}
\vspace{0.1cm}
\small
\setlength{\tabcolsep}{3.5pt}
\begin{tabular}{lrrrrrr}
\toprule
Method & NFE 1 & NFE 2 & NFE 4 & NFE 8 & NFE 10 & NFE 50 \\
\midrule
RF (backbone)
& 357.341 & 165.085 & 50.014 & 16.821 & 13.123 & 5.180 \\
\midrule
Teacher Reflow (MSE)
& 11.417 & \textbf{7.008} & \textbf{5.915}
& \textbf{5.539} & \textbf{5.502} & \textbf{5.511} \\
One-step distillation (MSE)
& \textbf{9.436} & 128.435 & 149.356
& 163.575 & 167.244 & 182.325 \\
\midrule
Fresh-independent FT
& 356.823 & 165.896 & 50.278 & 16.853 & 13.110 & 5.033 \\
OT-CFM-FT (\cite{tong2023improving})
& 232.657 & 90.181 & 28.572 & 13.338 & 11.269 & 5.137 \\
AlignFlow (\cite{kong2026alignflow})
& 164.919 & 58.269 & 23.050 & 12.861 & 11.142 & 5.175 \\
SD-FM (\cite{mousavi2026flow})
& 159.871 & 58.982 & 23.201 & 12.727 & 11.003 & 5.087 \\
Noise-ReCouple (cf.\ \citet{hui2025not})
& 132.991 & 47.773 & 22.852 & 16.109 & 14.956 & 10.313 \\
\rowcolor{orange!12} ReCouple ($B=N$)
& \textbf{102.110} & \textbf{41.520} & 23.001
& 16.817 & 15.658 & 10.429 \\
\rowcolor{orange!12} ReCouple
& 123.159 & 45.722 & 23.007
& 16.619 & 15.535 & 10.946 \\
\rowcolor{orange!12} ReCouple-AS
& 122.554 & 45.333 & \textbf{22.389}
& 16.141 & 15.047 & 10.518 \\
\rowcolor{orange!12} ReCouple+
& 187.926 & 68.357 & 24.130
& 12.656 & 10.936 & 5.216 \\
\rowcolor{orange!12} ReCouple++
& 187.080 & 67.905 & 23.797
& \textbf{12.292} & \textbf{10.527} & \textbf{4.781} \\
\bottomrule
\end{tabular}
\end{table}

\begin{table}[!h]
\centering
\caption{CIFAR-10 FID results with the OT-CFM pretrained
backbone. Same setting as Table~\ref{tab:cifar-current-baselines}.}
\label{tab:cifar-otcfm-baselines}
\vspace{0.1cm}
\small
\setlength{\tabcolsep}{3.5pt}
\begin{tabular}{lrrrrrr}
\toprule
Method & NFE 1 & NFE 2 & NFE 4 & NFE 8 & NFE 10 & NFE 50 \\
\midrule
OT-CFM (backbone)
& 230.321 & 86.640 & 27.312 & 12.709 & 10.723 & 4.902 \\
\midrule
Teacher Reflow (MSE)
& 11.196 & \textbf{6.848} & \textbf{5.705}
& \textbf{5.291} & \textbf{5.234} & \textbf{5.208} \\
One-step distillation (MSE)
& \textbf{9.087} & 137.009 & 159.789
& 177.066 & 181.694 & 201.234 \\
\midrule
Fresh-independent FT
& 351.241 & 155.466 & 47.210 & 15.845 & 12.286 & 4.634 \\
OT-CFM-FT (\cite{tong2023improving})
& 228.462 & 87.273 & 27.764 & 12.771 & 10.717 & 4.745 \\
AlignFlow (\cite{kong2026alignflow})
& 163.365 & 57.590 & 22.516 & 12.260 & 10.568 & 4.788 \\
SD-FM (\cite{mousavi2026flow})
& 157.251 & 57.958 & 22.743 & 12.243 & 10.546 & 4.779 \\
Noise-ReCouple (cf.\ \citet{hui2025not})
& 132.299 & 47.230 & 22.280 & 15.596 & 14.496 & 10.001 \\
 \rowcolor{orange!12} ReCouple ($B=N$)
& \textbf{101.005} & \textbf{40.877} & 22.354
& 16.265 & 15.112 & 10.082 \\
 \rowcolor{orange!12} ReCouple
& 121.547 & 45.018 & 22.208
& 15.957 & 14.895 & 10.451 \\
 \rowcolor{orange!12} ReCouple-AS
& 120.577 & 44.297 & \textbf{21.750}
& 15.480 & 14.431 & 10.085 \\
 \rowcolor{orange!12} ReCouple+
& 185.146 & 66.818 & 23.335
& 11.967 & 10.266 & 4.785 \\
 \rowcolor{orange!12} ReCouple++
& 184.021 & 65.635 & 22.878
& \textbf{11.535} & \textbf{9.826} & \textbf{4.342} \\
\bottomrule
\end{tabular}
\end{table}

We evaluate the ReCouple family on the CIFAR-10 generation task in detail. 
We compare to the following baselines:  (i) Fresh-independent FT: training on independent coupling,  (ii) OT-CFM-FT: solving a minibatch-OT during each training step, (iii) SDOT baselines \citep{kong2026alignflow} and \citep{mousavi2026flow}. 
We post-train ReCouple and the other similar baselines for 50k iterations using the same optimization setup during training.

Overall, ReCouple++ performs competitively at NFE-50 relative to similar baselines while providing a favorable balance across the various NFEs. ReCouple itself outperforms the noise injection strategy recently introduced by \citep{hui2025not} (Noise-ReCouple) at NFE 1-2.
Within the ReCouple family itself, ReCouple ($B=2048$) tracks and shows similar trend in FID performance across the NFEs with ReCouple ($B=N)$. Adding spatial regularization helps to boost the performance of both ReCouple and ReCouple+ across the NFEs.
App.~\ref{app_experimental} provides  ablation study on $B$, $R$ and $M$, as well as other details.

Notably, ReCouple+ and ReCouple++ lead to higher FID at the ultra-low NFE regime compared to ReCouple and ReCouple-AS, in agreement with the discussion in subsection \ref{subsec_recouple++}, but they shine in the higher NFE regime.
For a broader evaluation, we also include rollout-based training methods, which involve rolling out the teacher network using adaptive solver (using $\geq 100$ NFEs), and therefore do not compare fairly with the other baselines.
The one-step distillation method specializes on one-step generation and performs poorly in the higher NFE regime.

\subsection{ImageNet-256 Image Generation}
We further evaluate the ReCouple family at scale by post-training a pretrained
SiT-XL/2 backbone on ImageNet-256. We compare against the original backbone
and minibatch OT under a matched sampling protocol. All results are computed
from 50K generated samples using Euler integration and classifier-free guidance
with scale $4.0$. 

Table~\ref{tab:imagenet256} shows a clear NFE-dependent crossover.
SD-FM obtains the lowest FID at NFE 1--2, whereas ReCouple++ obtains
the lowest FID from NFE 5 onward. Within the ReCouple family, plain
ReCouple performs best at NFE 1--2, while ReCouple++ performs best
from NFE 5 onward. The results support our central observation:  coupling choices
that are strongest at 1-2 steps need not be strongest once the
sampler has sufficient numerical resolution, while spatial regularization
is particularly beneficial from NFE~5 onward.

\begin{table*}[t]
    \centering
    \caption{
        ImageNet-256 FID results using the pretrained SiT-XL/2 backbone.
        All metrics are computed using 50K generated samples with Euler
        integration. Lower is better; the best result at each NFE is shown
        in bold. ReCouple++ uses pooling factor $M=2$.
    }
    \label{tab:imagenet256}
    \small
    \begin{tabular}{lrrrrrr}
        \toprule
        Method
        & NFE 1 & NFE 2 & NFE 5
        & NFE 10 & NFE 25 & NFE 50 \\
        \midrule
        SiT-XL/2 FM
        & 275.340 & 99.078 & 10.172
        & 12.175 & 12.724 & 12.645 \\

        Minibatch OT (\cite{tong2023improving})
        & 253.866 & 90.371 & 7.845
        & 9.036 & 9.437 & 9.360 \\

        SD-FM \citep{mousavi2026flow}
        & {\textbf{239.573}} & \textbf{80.828} & 8.801
        & 10.329 & 10.828 & 10.855 \\

        \rowcolor{orange!12} ReCouple
        & 245.571 & 82.684 & 8.989
        & 10.637 & 11.150 & 11.164 \\

        \rowcolor{orange!12} ReCouple-AS
        & 246.547 & 88.086 & 7.275
        & 7.051 & 7.432 & 7.539 \\

        \rowcolor{orange!12} ReCouple++ ($M=2$)
        & 246.249 & 87.645 & \textbf{7.226}
        & \textbf{6.960} & \textbf{7.358} & \textbf{7.302} \\
        \bottomrule
    \end{tabular}
\end{table*}

\begin{figure*}[!t]
   \caption{Qualitative comparison on ImageNet-256  using pretrained SiT-XL/2 at NFE 25.
    Each row shows samples generated from the same initial noise and class condition using the SiT baseline, Minibatch OT, ReCouple, ReCouple-AS, and ReCouple++. This matched comparison highlights the differences in visual fidelity and semantic consistency induced by each sampling strategy.}
    \label{fig:imagenet_nfe25}
    \centering
    \includegraphics[trim={0 0 0 0.5cm}, clip, width=0.95\textwidth]{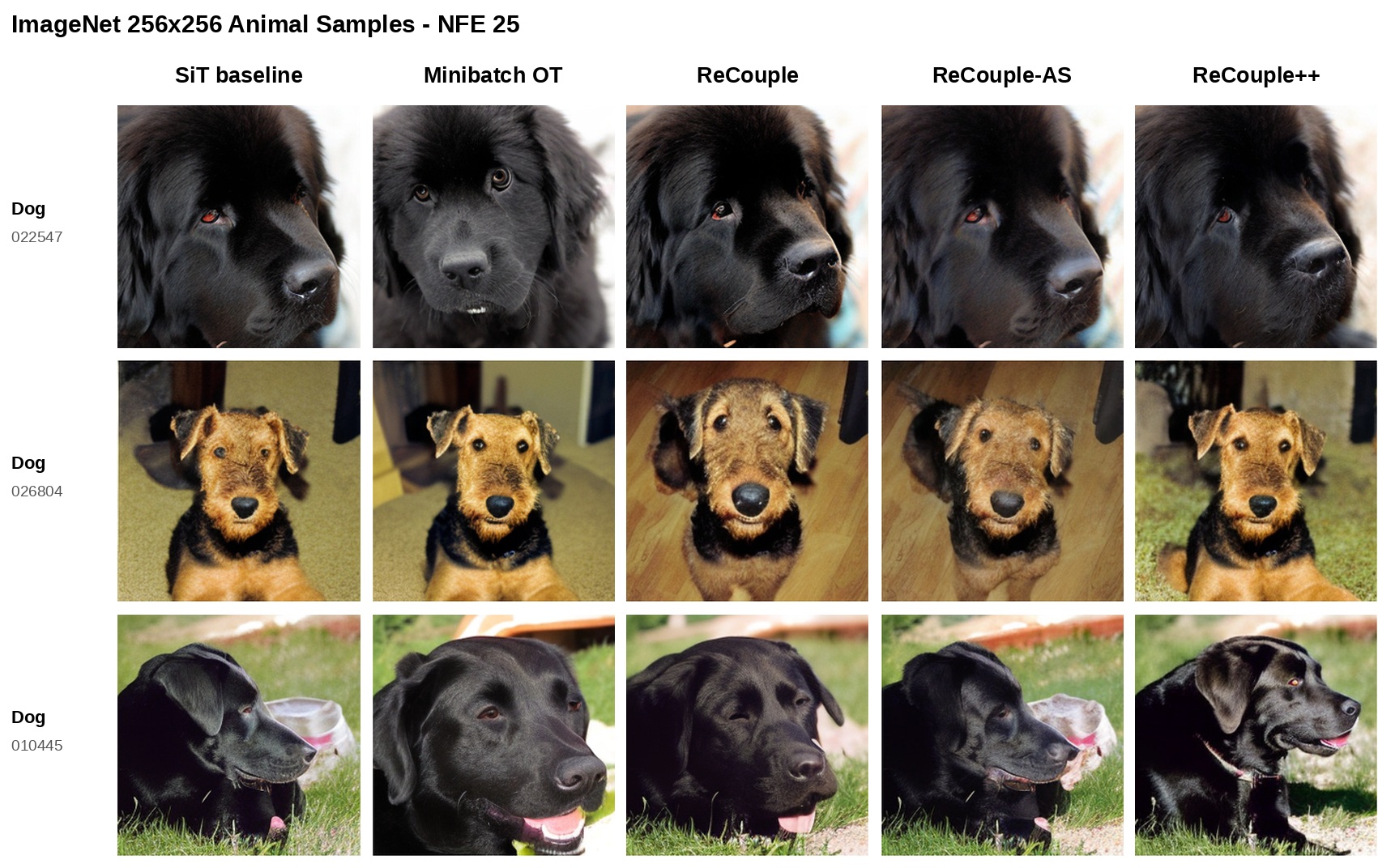}
\end{figure*}

\section{Conclusion}
We introduced the ReCouple family as a controllable framework for studying and designing endpoint couplings for post-training flow matching models, ranging from coarse blockwise transport to increasingly population-resolved and spatially regularized variants. 
Our findings  suggest putting less emphasis on approximating OT as closely as possible.  Instead,  coupling design  is better understood as a resolution matching problem. ReCouple controls the coarse-graining of the transport geometry, the pooling factor controls how densely that geometry is resolved from the continuous source, and  Jacobian regularization controls the complexity of  neural realization.

Our empirical results are consistent with this picture. ReCouple-based training gives competitive generative performance compared to similar baselines across low-NFE regime. Moreover, the ReCouple family gives an effective  yet controllable approach to understand generative improvement across  NFEs. 
The more interesting findings lie within the ReCouple family itself: refining the coupling improves transport fidelity, but may not lead to more substantial gains at the higher NFEs. Increasing source resolution favors higher NFEs at the expense of one-step performance, while spatial regularization mainly improves moderate-to-high NFEs.
Taken together, these results suggest that the most effective coupling need not be the most OT-like, but rather one whose transport resolution is compatible with the data support, the neural realization, and the numerical budget.

\section*{Acknowledgments}
The computations were enabled by resources provided by the National Academic Infrastructure for Supercomputing in Sweden (NAISS), funded by the Swedish Research Council (NAISS 2026/3-562).
SHL would like to acknowledge support from the Wallenberg Initiative on Networks and Quantum Information (WINQ) and the Swedish Research Council (VR/2021-03648).
MWM would like to acknowledge DARPA, NSF, the DOE Competitive Portfolios grant, and the DOE SciGPT grant for partial
support of this work.
NBE would like to acknowledge support from the U.S. Department of Energy, Office of Science, Office of Advanced Scientific Computing Research, through the EXPRESS: 2025 Exploratory Research for Extreme-Scale Science program and the RAPIDS Institute under the Scientific Discovery through Advanced Computing (SciDAC) program, under Contract No. DE-AC02-05CH11231 at Lawrence Berkeley National Laboratory. The views, opinions, and/or findings expressed are those of the authors and should not be interpreted as representing the official views or policies of the U.S. Government.

\bibliography{reference}
\bibliographystyle{plainnat}

\clearpage 

\appendix
\section*{Appendix}

\section{Related Work}
\label{app_relatedwork}

\paragraph{Flow matching (FM) and OT-guided flows} OT \citep{gabriel2019computational} defines  a natural geometry on the space of probability
measures and provides principled guidance to train  flow-based models:
OT-CFM and Multisample Flow Matching use minibatch OT, matching batches of noise samples to target samples, to reduce conflicting velocities and obtain straighter flows, improving sampling efficiency at low NFE \citep{tong2023improving,pooladian2023multisample}. However, they are not widely used in practice since the batch size needs to be large enough for them to perform well: the optimal matchings on small sample sizes are unstable and cannot
reliably approximate matches \citep{zhang2025fitting}. 
\cite{kornilov2024optimal, korotin2022neural, davtyan2025faster, chiang2026cot, ma2026learning}  propose alternative OT-based frameworks and techniques for improved few-step generation.  These works motivate OT as an important ingredient in efficient generation (see also \citep{lin2026arex} and related work). On the theory side, \citet{tsimpos2025eulerian, tsimpos2026one}  study characterizations, existence, and obstructions of straight flows in flow-based models, whereas \citet{lim2025hidden, li2026kinetic} study characterization and implicit bias via the lens of empirical FM (for sequential data, see \citep{lim2024elucidating, lim2026flow}). 

\paragraph{OT-guided flows versus neural realization} 
Despite the ideal nature of OT,  optimal couplings need not be easy for a neural network to realize. \cite{hui2025not} observes that straighter OT supervision can induce a more complex velocity field and introduces hybrid couplings to improve generative performance.
Model-Aligned Coupling (MAC)  instead asks which couplings the current model can learn most easily and selects pairs according to its prediction error \citep{lin2026beyond}. Similar to Reflow, MAC may become unreliable for post-training if the pretrained model is poor, since its coupling criterion is itself determined by the model’s predictions. These results distinguish transport optimality from neural approximation, which we leverage to extend ReCouple to Recouple-AS and ReCouple++ by additionally controlling spatial regularity of the learned velocity field.

\paragraph{Approximation of OT}
Large-batch Sinkhorn methods improve the approximation of OT but increase computational cost, while recent theory characterizes the population coupling induced by minibatch OT and its interaction with numerical integration \citep{boite2026expected}. Min-SWGG and sliced transport plans provide cheaper approximations through low-dimensional projections \citep{mahey2023fast}. A complementary line exploits the semi-discrete setting of generation, where the source distribution is continuous but only finitely many target samples are available: \cite{pooladian2023minimax} studies statistical properties, whereas SD-FM and AlignFlow propose efficient scheme to construct SDOT couplings between fresh source samples and the empirical data distribution \citep{mousavi2026flow,kong2026alignflow}. 
ReCouple is not primarily designed as a more accurate or efficient approximation to OT or SDOT. Instead, it introduces a controllable approximation to understand how coupling optimality affects the learnability and few-step behavior of FM models, complementing the recent line of  studies \citep{li2026kinetic, bertrand2026closed, briq2026amazing} that aim to provide a deeper understanding of flow-based models.

\paragraph{Rollout-free versus rollout-based post-training} Rectified Flow uses Reflow to progressively straighten model-induced couplings \citep{liu2022flow} but it requires first rolling out a teacher network before retraining. 
Moreover, Reflow is constrained by the quality of the pretrained flow model, which needs to be good enough to generate approximately the  target distribution, not to mention  the  cost needed to generate  a large number of data points at each rollout step. Another line of work accelerates pretrained diffusion or flow models through model generated trajectories \citep{lee2024improving, kim2025simple} or distillation \citep{sabour2026align}. Our framework does not use pretrained-model rollouts or teacher trajectories. Instead, it provides a rollout-free, controllable framework to study how coupling optimality, neural realizability, and low-NFE generation interact.

\clearpage 
\section{Mechanistic Insights on ReCouple} \label{app_recouple}

{\bf Why not just select random blocks?}
A natural question is whether the  blockwise procedure based on quantile sorting in ReCouple equivalent to selecting  random blocks, and if not, what are the advantages over the random baseline (RandomBlock\&Select). ReCouple and RandomBlock\&Select use the same sparsity budget but induce very different sparse transport graphs. For simplicity, assume that $B$ divides $N$ throughout this section. Any balanced block construction with block size $B$ retains \begin{equation} \frac{N}{B}B^2 = NB \end{equation} admissible source--target edges, corresponding to edge density $B/N$. Thus the difference between ReCouple and RandomBlock is not how many edges are retained, but which edges are retained.

Recall that in ReCouple, for a direction $\theta$, each sample is assigned its rank in the  ordering induced by the projection $x\mapsto\langle\theta,x\rangle$ after the sorting. ReCouple partitions these projected ranks into contiguous blocks of size $B$ and solves the full  assignment problem only within corresponding source and target rank blocks.
The projection in ReCouple is only 1D, but after data-dependent direction selection the induced sparse OT graph is very different from a uniformly random sparse graph. In particular, ReCouple searches for an ordering (induced by the best direction selected among the $R$ candidates) that makes the OT problem local and sparse. 

For comparison, define RandomBlock by conditioning on a balanced labeled source partition \begin{equation} S_1,\ldots,S_K, \qquad |S_k|=B, \end{equation} and drawing the labeled target partition \begin{equation} T_1,\ldots,T_K, \qquad |T_k|=B, \end{equation} uniformly from all balanced labeled target partitions. The corresponding block graph contains the edges \begin{equation} E_{\mathcal B} = \bigcup_{k=1}^K S_k\times T_k. \end{equation} 

\begin{proposition}[ReCouple versus random baseline] \label{prop:recouple-randomblock} 
Let \begin{equation} C^\star := \frac1N \min_{\pi\in S_N} \sum_{i=1}^N c_{i,\pi(i)} \end{equation} 
denote the unrestricted empirical quadratic OT cost. For a balanced block geometry $\mathcal B$, define \begin{equation} \mathcal F_{\mathcal B} := \left\{ \pi\in S_N: \pi(S_k)=T_k,\ \forall k \right\}, \end{equation} and \begin{equation} C_{\mathcal B} := \frac1N \min_{\pi\in\mathcal F_{\mathcal B}} \sum_{i=1}^N c_{i,\pi(i)}, \end{equation} 
where $c_{ij}:=\|x_i-y_j\|^2$. 
Then the following statements hold. 

\begin{enumerate}[label=(\roman*),leftmargin=2.2em] 
\item Every balanced block geometry of block size $B$ contains exactly $|E_{\mathcal B}|=NB$ admissible edges, and therefore has edge density $B/N$. 

\item Let $\pi^\star\in S_N$ be any fixed permutation and let \begin{equation} \operatorname{Rec}^{\rm rand}_B := \frac1N \sum_{i=1}^N \mathbf 1 \left\{ (i,\pi^\star(i)) \in E_{\mathcal B} \right\}. \end{equation} For RandomBlock, 
\begin{equation} \label{eq:random-recall-prop} \mathbb E[ \operatorname{Rec}^{\rm rand}_B ] = \frac{B}{N}. \end{equation} 
If $\pi^\star$ is the unique global OT permutation, then \begin{equation} \label{eq:random-exact-prop} \mathbb P( C_B^{\rm rand}=C^\star ) = \frac{(B!)^{N/B}}{N!}. \end{equation}

\item Fix a direction $\theta$. Let \begin{equation} \theta^\top x_{i_1} < \cdots < \theta^\top x_{i_N}, \qquad \theta^\top y_{j_1} < \cdots < \theta^\top y_{j_N}, \end{equation} and define the rank-matching permutation $\sigma_\theta(i_r)=j_r$. Let 
\begin{equation} C_{\rm rank}^\theta := \frac1N \sum_{r=1}^N \|x_{i_r}-y_{j_r}\|^2 \end{equation} denote the  quadratic cost of this rank matching and $C_B^\theta$ is the ReCouple blockwise cost induced by $\theta$. Then 
\begin{equation} \label{eq:rank-sandwich-prop} C^\star \leq C_B^\theta \leq C_{\rm rank}^\theta. \end{equation} In particular, \begin{equation} C_1^\theta=C_{\rm rank}^\theta, \qquad C_N^\theta=C^\star. 
\end{equation} 

\item Let $(u^\star,v^\star)$ be any optimal dual solution of the unrestricted assignment problem and define the  costs \begin{equation} \label{eq:reduced-cost-prop} r_{ij} := c_{ij}-u_i^\star-v_j^\star \geq0. \end{equation} Then, for any admissible block geometry $\mathcal B$, \begin{equation} \label{eq:restricted-reduced-cost} C_{\mathcal B}-C^\star = \frac1N \min_{\pi\in\mathcal F_{\mathcal B}} \sum_{i=1}^N r_{i,\pi(i)}. \end{equation} 
\end{enumerate}
\end{proposition}
Detailed proof is in App.~\ref{app_recouple-randomblock}.

Proposition~\ref{prop:recouple-randomblock} clarifies the advantage of ReCouple. The relevant criterion is not whether a sparse graph preserves every edge of one particular optimal permutation. By \eqref{eq:restricted-reduced-cost}, a sparse graph is sufficient whenever it supports \emph{some} perfect matching having small total reduced cost. OT-edge recall is therefore a useful geometric diagnostic, but it is not the optimization objective itself. ReCouple uses projected geometry to generate structured candidate graphs. In contrast, RandomBlock\&Select uses the same edge budget and the same best-of-$R$ selection principle, but its candidate balanced block graphs are generated without reference to the data geometry.

To separate the effect of the proposal law from the effect of best-of-$R$ selection, let 
\begin{equation} p_{\rm proj}(\varepsilon) := \mathbb P \left( \Delta_B^{\rm proj} \leq\varepsilon \right), \qquad p_{\rm RB}(\varepsilon) := \mathbb P \left( \Delta_B^{\rm RB} \leq\varepsilon \right), \end{equation} where $\Delta_B := C_B-C^\star$. 
For $R$ independent candidates from either proposal family, 
\begin{equation} \label{eq:best-R-proposal} \mathbb P \left( \min_{1\le r\le R}\Delta_B^{(r)} \leq\varepsilon \right) = 1- \left[ 1-p(\varepsilon) \right]^R. 
\end{equation} 

Hence the difference between ReCouple and RandomBlock\&Select is the lower-tail quality of their respective proposal distributions over sparse graphs. There is no universal result asserting $p_{\rm proj}(\varepsilon) \geq p_{\rm RB}(\varepsilon)$. Such an advantage is a property of the data geometry and can be measured empirically.

\begin{figure}[!h]
\caption{Sparse OT graph comparison at $B=8$ for ReCouple and RandomBlockSelect.}
    \centering
\includegraphics[width=0.96\linewidth]{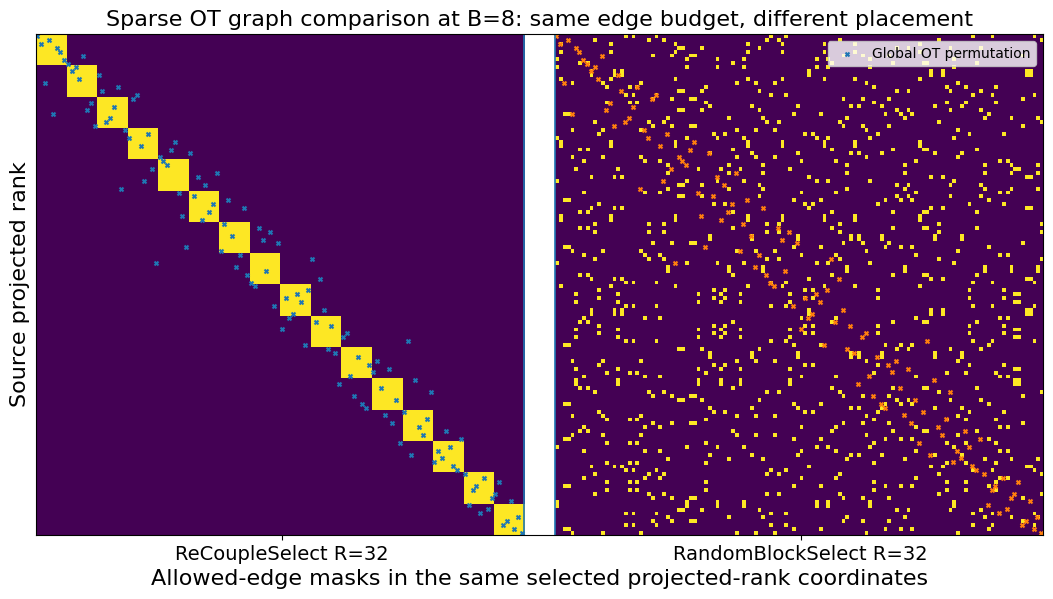}
    \label{fig:visual}
\end{figure}

Figure~\ref{fig:visual} illustrates this distinction in the selected ReCouple projected-rank coordinates for the case where the source samples are from the standard Gaussian and the target samples are from a 4-mode Gaussian. Both methods expose exactly $NB$ admissible edges. In this example, however, the direction selected by ReCouple makes the unrestricted empirical OT permutation noticeably more local in projected rank, so a larger fraction of its edges falls inside the diagonal $B\times B$ block graph. RandomBlock\&Select has the same sparsity and candidate budget but does not exploit this ordering.

When $B=N$, both methods coincide and are simply the global empirical OT. However, when $B<N$, they are not the same. ReCouple can approximate OT much faster and has a higher OT-edge recall at a small $B$ compared to the random baseline.

\begin{figure}[!h]
\caption{Gap to optimal coupling cost and OT-edge recall across $R$ at $B=8$. OT-edge recall is the fraction of edges in the global empirical OT matching
that are contained in the admissible block graph.}
    \centering
\includegraphics[scale=0.5]{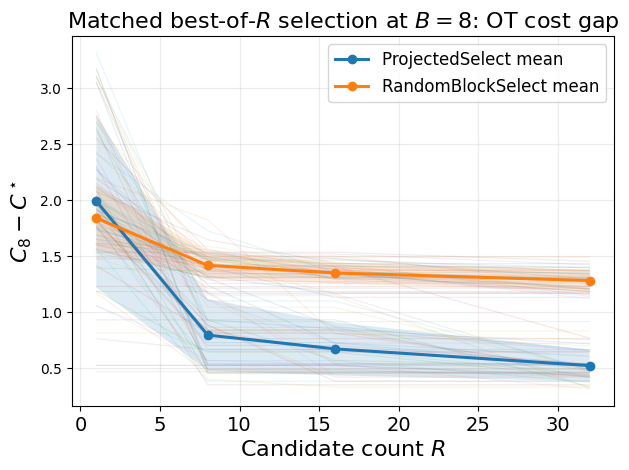} \includegraphics[scale=0.5]{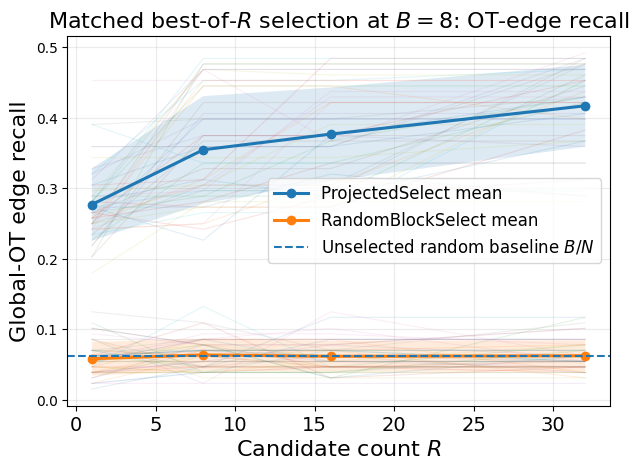}
\end{figure}

\begin{figure}[!h]
\caption{Held-out gap to optimal coupling cost and OT-edge recall across $B$ for various choices of $R$.}
    \centering
\includegraphics[scale=0.5]{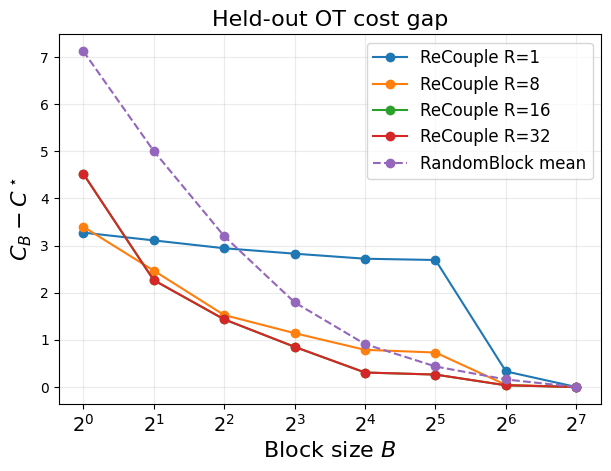}
\includegraphics[scale=0.5]{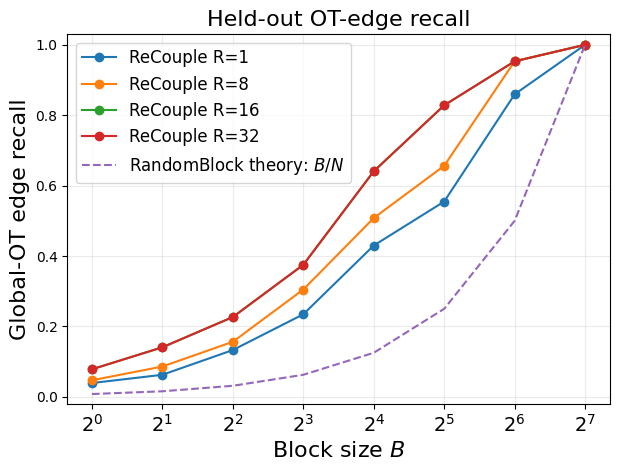}
\end{figure}

Table \ref{tab:matched-R-selection} shows that a single random projection is not necessarily better than a random block partition, but the best-of-$R$ selection is much more effective for ReCouple. At $B=8$, increasing $R$ from 1 to 32 reduces the ReCouple's gap from about 1.99 to 0.53, while RandomBlock\&Select only improves from about 1.84 to 1.28. Also, ReCouple's global-OT edge recall  rises to about 0.42, versus the random baseline $B/N=0.0625$. Overall these empirical results support the claim that ReCouple's best-of-$R$ search over structured space can find much better sparse OT graph than RandomBlock\&Select.

\begin{table}[t]
\centering
\caption{
Matched best-of-$R$ candidate selection for the standard Gaussian
source to four-mode Gaussian target experiment
($N=128$, $B=8$).
ReCouple chooses the best of $R$ random projection-induced block
geometries, while RandomBlock\&Select chooses the best of $R$ independent
geometry-blind balanced block geometries, using the same block-OT
objective.
Values are mean $\pm$ standard deviation over 50 independent candidate
pools.
The unrestricted global empirical OT cost is $C^\star$.
OT-edge recall is the fraction of edges of the global OT permutation
retained by the selected block graph; the geometry-blind baseline is
$B/N=0.0625$.
}
\label{tab:matched-R-selection}
\vspace{0.15cm}
\small
\setlength{\tabcolsep}{4.5pt}
\begin{tabular}{c
                rr
                rr}
\toprule
& \multicolumn{2}{c}{$C_8-C^\star$ $\downarrow$}
& \multicolumn{2}{c}{OT-edge recall $\uparrow$} \\
\cmidrule(lr){2-3}
\cmidrule(lr){4-5}
$R$
& ReCouple
& RandomBlock\&Select
& ReCouple
& RandomBlock\&Select \\
\midrule
1
& $1.991 \pm 0.772$
& $\mathbf{1.840 \pm 0.292}$
& $\mathbf{0.277 \pm 0.051}$
& $0.058 \pm 0.024$ \\

8
& $\mathbf{0.797 \pm 0.319}$
& $1.418 \pm 0.114$
& $\mathbf{0.355 \pm 0.076}$
& $0.064 \pm 0.023$ \\

16
& $\mathbf{0.674 \pm 0.236}$
& $1.351 \pm 0.094$
& $\mathbf{0.377 \pm 0.066}$
& $0.062 \pm 0.022$ \\

32
& $\mathbf{0.525 \pm 0.141}$
& $1.283 \pm 0.094$
& $\mathbf{0.417 \pm 0.058}$
& $0.063 \pm 0.022$ \\
\bottomrule
\end{tabular}
\end{table}

{\bf Why use the 1D random slices/projections?} 
ReCouple has the broad structure of ``partition, solve OT locally, and
combine,'' but its blocks are not clusters in the usual Euclidean or
semantic sense. A unit direction $\theta\in\mathbb \mathcal{S}^{d-1}$ defines the
shared scalar score
$s_\theta(x)=\langle\theta,x\rangle$
and the source and target samples are partitioned into corresponding
equal-size intervals of their projected ranks.
The projection is used only to construct the admissible graph.

Other shared scalar scores could in principle be used to construct
equal-rank blocks. Linear projections are particularly attractive because
they are inexpensive, require no additional metric or bandwidth, and have
a direct relation to Euclidean quadratic cost:
\begin{equation}
    |\langle\theta,x-y\rangle|
    \leq
    \|x-y\|,
    \qquad
    \mathbb E_{\theta\sim{\rm Unif}(\mathbb S^{d-1})}
    |\langle\theta,x-y\rangle|^2
    =
    \frac1d\|x-y\|^2.
\end{equation}

Finally, ReCouple does not rely on a single projection. It searches over
several candidate directions and selects the one having the
lowest full-dimensional blockwise OT cost. Its advantage over a purely
random block construction is therefore not guaranteed by
projection alone; it is more pronounced when the projection proposal family places
more probability on  graphs supporting low-cost perfect matchings.

{\bf Relation with SWGG \citep{mahey2023fast}.} The core idea of ReCouple is similar in spirit to SWGG: both involves taking the rank matching induced by a random slice/projection and lifts that permutation back to the original $d$-dimensional samples.
However, SWGG uses it to define a distributional discrepancy, whereas ReCouple uses it to define  structured feasible sets for OT approximation.

To be precise, let
$\widehat\mu=\frac1N\sum_{i=1}^N\delta_{y_i},$
$\widehat\nu=\frac1N\sum_{i=1}^N\delta_{z_i},$
and, for $\theta\in\mathbb S^{d-1}$, let
$\sigma_y^\theta,\sigma_z^\theta\in S_N$ sort the projected samples
$\langle\theta,y_i\rangle$ and $\langle\theta,z_i\rangle$.
The induced rank coupling is
$\pi_\theta\!\left(\sigma_y^\theta(k)\right) = \sigma_z^\theta(k),$
$k=1,\ldots,N.$
For quadratic cost,
$C_N(\pi) := \frac1N\sum_{i=1}^N\|y_i-z_{\pi(i)}\|^2,$
the fixed-direction SWGG cost is precisely
$\operatorname{SWGG}_2^2(\widehat\mu,\widehat\nu;\theta) = C_N(\pi_\theta),$
and
$\operatorname{min\text{-}SWGG}_2^2(\widehat\mu,\widehat\nu) = \min_{\theta\in\mathbb S^{d-1}}C_N(\pi_\theta)$
\citep{mahey2023fast}.

For ReCouple, let $\Pi_B(\theta)\subseteq S_N$ denote the permutations
that match only within corresponding blocks of $B$ consecutive
projected ranks, and define
$\pi_{\theta,B} \in \arg\min_{\pi\in\Pi_B(\theta)}C_N(\pi).$
Given directions $\theta_1,\ldots,\theta_R$, ReCouple selects
$\widehat\pi_{B,R} \in \arg\min_{1\le r\le R}C_N(\pi_{\theta_r,B}),$
up to the additional identity candidate used in our implementation.
\begin{itemize}
\item When $B=1$,
$\Pi_1(\theta)=\{\pi_\theta\},$
and therefore
$\widehat\pi_{1,R} \in \arg\min_{1\le r\le R} \operatorname{SWGG}_2^2(\widehat\mu,\widehat\nu;\theta_r).$
Thus ReCouple with $B=1$ is exactly a finite best-of-$R$ search
over the projection-induced transport plans underlying min-SWGG
(with the identity candidate providing one additional option).
\item 
For $B>1$, the SWGG rank coupling remains feasible,
$\pi_\theta\in\Pi_B(\theta)$, hence
$C_N(\pi_{\theta,B}) \le C_N(\pi_\theta) = \operatorname{SWGG}_2^2(\widehat\mu,\widehat\nu;\theta).$
Consequently,
$$W_2^2(\widehat\mu,\widehat\nu) \le C_N(\widehat\pi_{B,R}) \le \min_{1\le r\le R} \operatorname{SWGG}_2^2(\widehat\mu,\widehat\nu;\theta_r),$$
where inclusion of the identity candidate can only further decrease
the middle quantity. On the other hand, 
$B=N \implies \Pi_N(\theta)=S_N,$
so the inner problem reduces to empirical global OT.

\end{itemize}

Therefore, ReCouple interpolates between the finite-search min-SWGG plans ($B=1$) and the global empirical OT ($B=N)$.

This connection is to SWGG rather than ordinary sliced Wasserstein.
Sliced Wasserstein evaluates the 1D projected transport
cost, whereas SWGG lifts the projection-induced rank pairing back to
the ambient space and evaluates
$\frac1N\sum_{k=1}^N \left\| y_{\sigma_y^\theta(k)} - z_{\sigma_z^\theta(k)} \right\|^2.$
ReCouple uses this same projected rank structure to define a
feasible set for the original $d$-dimensional assignment
problem. 

We end with a caveat. Note that the construction of \cite{mahey2023fast}  is called min-SW lifted transport plans in \citep{peyre2025optimal} (see Chapter 11.2 there), which exposes that min-SW does not metrize weak convergence. However, the projection is not the transport approximation itself. It is merely a cheap mechanism for constructing a structured sparse support on which one then solves the original ambient assignment problem (see also Figure 11.6 in \citep{peyre2025optimal}). ReCouple is an extension which shares the same principle: 
it samples $R$ 1D projections to generate candidate coarse rank partitions, solves the exact ambient-space assignment within blocks of size $B$, and selects the lowest-cost candidate, with $B$ controlling the locality of the transport constraint and $R$ controlling the search over admissible coarse-grainings.

\section{Detailed Algorithms}
\label{app_detailedalgo}

In this section, we provide detailed algorithm for the post-training scheme, as well as ReCouple, ReCouple+ and ReCouple++.

\begin{algorithm}[!h]
\caption{Vanilla post-training on a fixed  coupling}
\label{alg:main_posttrain-coupling}
\begin{algorithmic}[1]
\Require fixed coupling $\widehat\gamma$; pretrained backbone $v^{w_0}$ with parameters $w_0$;
EMA rate $\beta_{\rm EMA}$; post-training budget $J$.
\State Initialize $w\leftarrow w_0$ and $\bar w\leftarrow w_0$.
\For{$j=1,\ldots,J$}
    \State Draw  $(Z,Y)\sim\widehat\gamma$ and $t\sim \mathrm{Unif}[0,1]$.
    \State Set $X_t=(1-t)Z+tY$ and $U=Y-Z$.
    \State Update $w$ by one optimizer step on
    $\|v_t^w(X_t)-U\|^2$.
    \State Update the EMA parameters
    $\bar w\leftarrow\beta_{\rm EMA}\bar w+(1-\beta_{\rm EMA})w$.
\EndFor
\State \Return the velocity field $v^{\bar w}$.
\end{algorithmic}
\end{algorithm}

Algorithm~\ref{alg:recouple-main} assumes $B$ divides $N$ for simplicity.
The following general form allows a possibly smaller final projected-rank block
when $B$ does not divide $N$ and reduces to
Algorithm~\ref{alg:recouple-main} when $B$ divides $N$.

\begin{algorithm}[!t]
\caption{ReCouple: general case}
\label{alg:recouple-general}
\begin{algorithmic}[1]
\Require target bank $y_{1:N}$; source bank $z_{1:N}$;
block size $B\leq N$; proposal count $R$

\State Add the identity pairing $\pi_0(i)=i$, $i=1,\ldots,N$, to the candidate set.

\State Set
\[
Q=\left\lceil \frac{N}{B}\right\rceil,
\qquad
I_b
=
\{(b-1)B+1,\ldots,\min\{bB,N\}\},
\qquad b=1,\ldots,Q.
\]

\For{$r=1,\ldots,R$}
    \State Draw
    $\theta_r\sim\mathrm{Unif}(\mathbb S^{d-1})$.

    \State Let $\sigma_r,\tau_r\in S_N$ sort
    $\{\langle\theta_r,y_i\rangle\}_{i=1}^N$ and
    $\{\langle\theta_r,z_j\rangle\}_{j=1}^N$, respectively.

    \For{$b=1,\ldots,Q$}
        \State Compute
        \[
        \rho_{r,b}
        \in
        \arg\min_{\rho\in\mathfrak S(I_b)}
        \sum_{k\in I_b}
        \left\|
        y_{\sigma_r(k)}
        -
        z_{\tau_r(\rho(k))}
        \right\|^2.
        \]
    \EndFor

    \State Define $\pi_r\in S_N$ by
    \[
    \pi_r(\sigma_r(k))
    =
    \tau_r(\rho_{r,b}(k)),
    \qquad
    k\in I_b,\quad b=1,\ldots,Q.
    \]

    \State Evaluate
    \[
    C_N(\pi_r)
    =
    \frac{1}{N}\sum_{i=1}^N
    \|y_i-z_{\pi_r(i)}\|^2.
    \]
\EndFor

\State Evaluate
$C_N(\pi_0)
=
\frac{1}{N}\sum_{i=1}^N
\|y_i-z_{\pi_0(i)}\|^2.$

\State Select
$r^\star
\in
\arg\min_{r=0,\ldots,R}
C_N(\pi_r),
\  \
\widehat\pi=\pi_{r^\star}.$

\State \Return
$\widehat\gamma
=
\frac{1}{N}
\sum_{i=1}^N
\delta_{(z_{\widehat\pi(i)},\,y_i)}.$

\end{algorithmic}
\end{algorithm}

\begin{algorithm}[!h]
\caption{ReCouple+}
\label{alg:recouple-plus}
\begin{algorithmic}[1]
\Require target bank $y_{1:N}$; source sampler $Z\sim\nu$; designated source bank $z^{(1)}_{1:N}$; block size $B \leq N$; proposal count $R$; pooling factor $M$.

\State Run the proposal stage of Algorithm~\ref{alg:recouple-main} on $(y_{1:N},z^{(1)}_{1:N})$ and select

$$
\theta^\star=\theta_{r^\star},
\qquad
r^\star\in\arg\min_{1\le r\le R} C_N(\pi_r).
$$

\State Form a pooled source bank $z^{(M)}_{1:MN}$ whose first $N$ samples are $z^{(1)}_{1:N}$ and whose remaining samples are independent draws from $\nu$.

\State Order the target and pooled source banks by projection onto $\theta^\star$, and relabel them in this order.
\State Partition the ordered target bank into contiguous blocks $I_q$ of size $b_q\le B$ and associate each $I_q$ with the corresponding block $J_q^{(M)}$ of $Mb_q$ pooled sources.

\For{each block $q$}
\State Solve the capacity-$M$ assignment problem (discrete Kantorovich):

$$
\min_{A_{ij}\in\{0,1\}}
\sum_{i\in I_q}\sum_{j\in J_q^{(M)}}
A_{ij}\|y_i-z_j\|^2,
$$

subject to
$
\sum_{i\in I_q}A_{ij}=1$ and
$\sum_{j\in J_q^{(M)}}A_{ij}=M.
$

\EndFor

\State Concatenate the block solutions to obtain the  pooled permutation
$\widehat\pi^+$ between the $MN$ duplicated target slots and the $MN$ pooled source samples. \\
\Return $\widehat{\gamma}^{+}
=
\frac{1}{MN}
\sum_{i=1}^{N}
\sum_{m=1}^{M}
\delta_{\left(
z^{(M)}_{\widehat{\pi}^{+}(i,m)},
\,y_i
\right)}.$
\end{algorithmic}
\end{algorithm}

\begin{algorithm}[!t]
\caption{Post-training on a fixed coupling with annealed spatial Jacobian regularization}
\label{alg:posttrain-coupling}
\begin{algorithmic}[1]
\Require fixed coupling $\widehat\gamma$; pretrained backbone $v^{w_0}$;
EMA rate $\beta_{\rm EMA}$; post-training budget $J$;
tuning parameters $\sigma,\beta_{\rm reg},\rho$.

\State Initialize $w\leftarrow w_0$, $\bar w\leftarrow w_0$, and $J_{\rm AS}=\max\{2,\operatorname{round}(\rho J)\}$.
\State If $\beta_{\rm reg}>0$, on a fixed calibration batch evaluate the initial CFM loss $L_0$ and AS loss $R_0$ under $w_0$, and set $\lambda_0=\beta_{\rm reg}L_0/R_0$; otherwise set $\lambda_0=0$.

\For{$j=0,\ldots,J-1$}
    \State Draw $(Z,Y)\sim\widehat\gamma$ and
    $t\sim\mathrm{Unif}[0,1]$.
    \State Set $X_t=(1-t)Z+tY$ and $U=Y-Z$.
    \State Set
    $\mathcal L_{\rm CFM}=\|v_t^w(X_t)-U\|^2$.
    \State Set
    $\lambda_j=\frac{\lambda_0}{2}
    [1+\cos(\pi j/J_{\rm AS})]$ if $j<J_{\rm AS}$,
    and $\lambda_j=0$ otherwise.
    \If{$\lambda_j>0$}
        \State Draw $\xi\sim\mathcal N(0,I_d)$ and set
        $\mathcal R_\sigma=
        (2\sigma^2)^{-1}
        \|v_t^w(X_t+\sigma\xi)-v_t^w(X_t)\|^2$.
    \Else
        \State Set $\mathcal R_\sigma=0$.
    \EndIf
    \State Update $w$ by one optimizer step on
    $\mathcal L_{\rm CFM}+\lambda_j\mathcal R_\sigma$.
    \State Update
    $\bar w\leftarrow
    \beta_{\rm EMA}\bar w+(1-\beta_{\rm EMA})w$.
\EndFor
\State \Return the EMA velocity field $v^{\bar w}$.
\end{algorithmic}
\end{algorithm}

\clearpage
\section{Theoretical Results and Proofs} \label{app_proof}

Recall that the squared 2-Wasserstein distance is defined as:
$$W_2^2(\nu,\mu) :=
\inf_{\gamma\in\Pi(\nu,\mu)}
\int\| x-y\|^2\, \mathrm{d} \gamma(x,y) = \inf_{\gamma\in\Pi(\nu,\mu)} \mathbb{E}_\gamma \| X - Y\|^2.$$

\subsection{Proof of Proposition \ref{prop:sdot-straight}}

\begin{proof} 
Since $T=\nabla\phi$ for a convex potential $\phi$ by Brenier's theorem \citep{villani2021topics}, $T$ is monotone: \begin{equation} 
\langle T(z)-T(z'),z-z'\rangle\geq 0.  
\end{equation}

For $t<1$, the interpolation map \begin{equation} T_t=(1-t)\mathrm{Id}+tT \end{equation} is therefore strongly monotone, since \begin{equation} \begin{split} \langle T_t(z)-T_t(z'),z-z'\rangle &= (1-t)\|z-z'\|^2 + t\langle T(z)-T(z'),z-z'\rangle\\ &\geq (1-t)\|z-z'\|^2. \end{split} \end{equation} Hence $T_t$ is injective for every $t<1$. Consequently, $Z$ is a measurable function of $X_t=T_t(Z)$, and therefore so is $\dot X_t=T(Z)-Z$. It follows that \begin{equation} \operatorname{Cov}(\dot X_t\mid X_t)=0 \end{equation} and \begin{equation} v_t^*(X_t) = \mathbb E[\dot X_t\mid X_t] = \dot X_t. \end{equation} Since $\dot X_t=T(Z)-Z$ is independent of $t$ along each characteristic, its material acceleration vanishes. \end{proof}

\subsection{Proof of a Detailed Version of Theorem \ref{thm:fixed-N-SDOT} and Its Corollary}

The following is a detailed version of Theorem \ref{thm:fixed-N-SDOT}.

\begin{theorem}[Global-block ReCouple+ converges to SDOT]
\label{thm:fixed-N-SDOT_detailed}
Let $\widehat\mu_N
    = \frac1N\sum_{i=1}^N\delta_{y_i}$
be fixed, and let
$\nu\in\mathcal P_2(\mathbb R^d)$ be absolutely continuous.
Let $(Z_j)_{j\geq1}$ be i.i.d.\ with law $\nu$, and define
$ \widehat\nu_{MN}
    =
    \frac{1}{MN}
    \sum_{j=1}^{MN}\delta_{Z_j}.$
Consider ReCouple+ with a single global block, $B=N$, and pooling factor
$M$. Let $\widehat\gamma_{N,M}^{+}$ be an optimal coupling produced by the
resulting capacitated assignment, and let
\begin{equation}
    \widehat C_{N,M}^{+}
    =
    \int_{\mathbb R^d\times\mathbb R^d}
    \|z-y\|^2
    \,d\widehat\gamma_{N,M}^{+}(z,y).
\end{equation}

Then there exists an event $\Omega_0$ with
$\mathbb P(\Omega_0)=1$ such that, for every $\omega\in\Omega_0$ and for
every choice of finite-$M$ optimal couplings
$\widehat\gamma_{N,M}^{+}(\omega)$,
\begin{equation}
\label{eq:cost-sdot-limit}
    \widehat C_{N,M}^{+}
    \longrightarrow
    W_2^2(\nu,\widehat\mu_N)
    \qquad\text{as }M\to\infty,
\end{equation}
and
\begin{equation}
\label{eq:plan-sdot-limit}
    W_2
    \left(
        \widehat\gamma_{N,M}^{+},
        \gamma_N^{\rm SDOT}
    \right)
    \longrightarrow0,
\end{equation}
where $\gamma_N^{\rm SDOT}$ is the unique quadratic optimal coupling
between $\nu$ and $\widehat\mu_N$.
\end{theorem}

\begin{proof}
We proceed in seven steps.

\medskip
\noindent
\textbf{Step 1: For $B=N$, ReCouple+ coincides with empirical OT.}

When $B=N$, all $N$ target samples lie in one block. Hence the projection
direction and projected ordering impose no restriction: every pooled source
sample may be assigned to every target sample.

Replicate each target $M$ times and introduce the labeled target slots
\begin{equation}
    y_{i,r}=y_i,
    \qquad
    i=1,\ldots,N,
    \quad
    r=1,\ldots,M.
\end{equation}
Each of the $MN$ source samples and each of the $MN$ target slots has mass
$1/(MN)$. After scaling by $MN$, a transport plan between these two
equal-weight empirical measures is represented by a doubly stochastic
$MN\times MN$ matrix.

By the Birkhoff-von Neumann theorem (see, e.g., Theorem 3.17 in \citep{peyre2025optimal}), every doubly stochastic matrix is a
convex combination of permutation matrices. Since the transport objective
is linear, an optimum is attained at a permutation matrix. Therefore the
optimal transport problem between the pooled source samples and the replicated
target slots is exactly the assignment problem solved by global-block
ReCouple+.

It remains to relate the replicated target problem to transport toward
$\widehat\mu_N$. Collapsing the $M$ copies of each $y_i$ maps any feasible
coupling for the replicated problem to a coupling in
\begin{equation}
    \Pi(
        \widehat\nu_{MN},
        \widehat\mu_N
    )
\end{equation}
without changing its cost. Conversely, any coupling in
$\Pi(\widehat\nu_{MN},\widehat\mu_N)$ can be lifted to the replicated
problem by splitting the total mass assigned to each $y_i$ among its $M$
identical slots; again, the quadratic cost is unchanged. Hence the two
transport problems have the same optimal value.

Consequently,
\begin{equation}
\label{eq:empirical-ot-exact}
    \widehat C_{N,M}^{+}
    =
    W_2^2(
        \widehat\nu_{MN},
        \widehat\mu_N
    ),
\end{equation}
and, after collapsing the replicated target slots,
\begin{equation}
\label{eq:finite-M-opt}
    \widehat\gamma_{N,M}^{+}
    \in
    \Pi_{\rm opt}
    (
        \widehat\nu_{MN},
        \widehat\mu_N
    ).
\end{equation}
Thus, for $B=N$, ReCouple+ solves the finite-source empirical OT problem
exactly.

\medskip
\noindent
\textbf{Step 2: Almost-sure $W_2$ convergence of the empirical source.}

Since $\nu\in\mathcal P_2(\mathbb R^d)$, there exists an event
$\Omega_0$ of probability one on which both
\begin{equation}
    \widehat\nu_n
    \rightharpoonup
    \nu
\end{equation}
and
\begin{equation}
    \frac1n\sum_{j=1}^n\|Z_j\|^2
    \longrightarrow
    \int\|z\|^2\,d\nu(z)
\end{equation}
hold as $n\to\infty$. We henceforth fix an arbitrary
$\omega\in\Omega_0$ and argue deterministically.

Restricting to the subsequence $n=MN$ gives
\begin{equation}
    \widehat\nu_{MN}
    \rightharpoonup
    \nu
\end{equation}
and
\begin{equation}
    \int\|z\|^2\,d\widehat\nu_{MN}(z)
    \longrightarrow
    \int\|z\|^2\,d\nu(z).
\end{equation}
Weak convergence together with convergence of second moments is equivalent
to convergence in $W_2$, hence
\begin{equation}
\label{eq:source-W2}
    W_2(
        \widehat\nu_{MN},
        \nu
    )
    \longrightarrow0.
\end{equation}

\medskip
\noindent
\textbf{Step 3: Convergence of the optimal costs.}

Since $\widehat\mu_N$ is fixed, the triangle inequality gives
\begin{equation}
\begin{split}
    \bigl|
        W_2(
            \widehat\nu_{MN},
            \widehat\mu_N
        )
        -
        W_2(
            \nu,
            \widehat\mu_N
        )
    \bigr|
    &\leq
    W_2(
        \widehat\nu_{MN},
        \nu
    ).
\end{split}
\end{equation}
By \eqref{eq:source-W2},
\begin{equation}
    W_2(
        \widehat\nu_{MN},
        \widehat\mu_N
    )
    \longrightarrow
    W_2(
        \nu,
        \widehat\mu_N
    ).
\end{equation}
Together with \eqref{eq:empirical-ot-exact}, this yields
\begin{equation}
    \widehat C_{N,M}^{+}
    \longrightarrow
    W_2^2(
        \nu,
        \widehat\mu_N
    ).
\end{equation}

\medskip
\noindent
\textbf{Step 4: Tightness of the empirical optimal couplings.}

The first and second marginals of
$\widehat\gamma_{N,M}^{+}$ are respectively
$\widehat\nu_{MN}$ and $\widehat\mu_N$. Therefore
\begin{equation}
\begin{split}
    \int
    \bigl(
        \|z\|^2+\|y\|^2
    \bigr)
    \,d\widehat\gamma_{N,M}^{+}(z,y)
    &=
    \int\|z\|^2
    \,d\widehat\nu_{MN}(z)
    +
    \int\|y\|^2
    \,d\widehat\mu_N(y).
\end{split}
\end{equation}
The right-hand side is uniformly bounded in $M$. Hence the sequence
$\{\widehat\gamma_{N,M}^{+}\}_{M\geq1}$ is tight.

By Prokhorov's theorem, every subsequence
$\{\widehat\gamma_{N,M_k}^{+}\}$ admits a further subsequence
$\{\widehat\gamma_{N,M_{k_\ell}}^{+}\}$ such that
\begin{equation}
\label{eq:subseq-limit}
    \widehat\gamma_{N,M_{k_\ell}}^{+}
    \rightharpoonup
    \bar\gamma
\end{equation}
for some probability measure $\bar\gamma$ on
$\mathbb R^d\times\mathbb R^d$.

\medskip
\noindent
\textbf{Step 5: Every subsequential limit has the correct marginals.}

The coordinate projections are continuous. Hence, taking marginals in
\eqref{eq:subseq-limit},
\begin{equation}
    (\operatorname{proj}_1)_\#
    \bar\gamma
    =
    \nu,
    \qquad
    (\operatorname{proj}_2)_\#
    \bar\gamma
    =
    \widehat\mu_N.
\end{equation}
Thus
\begin{equation}
    \bar\gamma
    \in
    \Pi(
        \nu,
        \widehat\mu_N
    ).
\end{equation}

\medskip
\noindent
\textbf{Step 6: Every subsequential limit is the SDOT coupling.}

The quadratic cost
\begin{equation}
    c(z,y)=\|z-y\|^2
\end{equation}
is nonnegative and lower semicontinuous. Therefore
\begin{equation}
\begin{split}
    \int
    \|z-y\|^2
    \,d\bar\gamma(z,y)
    &\leq
    \liminf_{\ell\to\infty}
    \int
    \|z-y\|^2
    \,d\widehat\gamma_{N,M_{k_\ell}}^{+}(z,y)\\
    &=
    W_2^2(
        \nu,
        \widehat\mu_N
    ),
\end{split}
\end{equation}
where Step~3 was used in the last equality.

Since $\bar\gamma\in\Pi(\nu,\widehat\mu_N)$,
\begin{equation}
    \int
    \|z-y\|^2
    \,d\bar\gamma(z,y)
    \geq
    W_2^2(
        \nu,
        \widehat\mu_N
    ).
\end{equation}
Hence equality holds, and $\bar\gamma$ is an optimal coupling between
$\nu$ and $\widehat\mu_N$.

Because $\nu$ is absolutely continuous, Brenier's theorem implies that the
quadratic optimal coupling is unique. Therefore
\begin{equation}
    \bar\gamma
    =
    \gamma_N^{\rm SDOT}.
\end{equation}

We have shown that every subsequence admits a further subsequence converging
weakly to the same limit $\gamma_N^{\rm SDOT}$. Consequently,
\begin{equation}
\label{eq:weak-whole}
    \widehat\gamma_{N,M}^{+}
    \rightharpoonup
    \gamma_N^{\rm SDOT}
    \qquad
    \text{as }M\to\infty.
\end{equation}

\medskip
\noindent
\textbf{Step 7: Upgrade weak convergence to $W_2$ convergence.}

With the product Euclidean norm on
$\mathbb R^d\times\mathbb R^d$,
\begin{equation}
    \|(z,y)\|^2
    =
    \|z\|^2+\|y\|^2.
\end{equation}
Hence
\begin{equation}
\begin{split}
    &\int
    \|(z,y)\|^2
    \,d\widehat\gamma_{N,M}^{+}(z,y)\ = 
    \int\|z\|^2\,d\widehat\nu_{MN}(z)
    +
    \int\|y\|^2\,d\widehat\mu_N(y)\\
    &\qquad\longrightarrow
    \int\|z\|^2\,d\nu(z)
    +
    \int\|y\|^2\,d\widehat\mu_N(y) =
    \int
    \|(z,y)\|^2
    \,d\gamma_N^{\rm SDOT}(z,y).
\end{split}
\end{equation}
Combining this second-moment convergence with
\eqref{eq:weak-whole} yields
\begin{equation}
    W_2
    \left(
        \widehat\gamma_{N,M}^{+},
        \gamma_N^{\rm SDOT}
    \right)
    \longrightarrow0.
\end{equation}
Since the argument holds for every $\omega\in\Omega_0$ and for every
choice of finite-$M$ optimal couplings, the result follows almost surely.
\end{proof}

ReCouple+ converges to SDOT in the joint limit of $B \to N$ and $M \to \infty$.

\begin{corollary}[Joint limit of ReCouple+]
\label{cor:joint-BM-SDOT}
Fix $N$. Under the assumptions of
Theorem~\ref{thm:fixed-N-SDOT}, let $(B_k,M_k)$ satisfy
\begin{equation}
    B_k\in\{1,\ldots,N\},
    \qquad
    B_k\to N,
    \qquad
    M_k\to\infty .
\end{equation}
Then there exists an event $\Omega_0$ with
$\mathbb P(\Omega_0)=1$ such that, on $\Omega_0$, for every such sequence
$(B_k,M_k)$ and every choice of optimal ReCouple+ couplings,
\begin{equation}
    \widehat C_{N,B_k,M_k}^{+}
    \longrightarrow
    W_2^2(\nu,\widehat\mu_N),
\end{equation}
and
\begin{equation}
    W_2\!\left(
        \widehat\gamma_{N,B_k,M_k}^{+},
        \gamma_N^{\rm SDOT}
    \right)
    \longrightarrow0 .
\end{equation}
\end{corollary}

\begin{proof}[Proof of Corollary~\ref{cor:joint-BM-SDOT}]
Because $B_k\in\{1,\ldots,N\}$ and $B_k\to N$, there exists
$k_0$ such that
$B_k=N$, for all $k\geq k_0.$
Hence, after discarding finitely many terms, ReCouple+ has a single global
block and the projected-rank restriction is absent.

By Theorem~\ref{thm:fixed-N-SDOT}, there exists an event
$\Omega_0$ of probability one on which, for every choice of finite-$M$
optimal couplings,
\begin{equation}
    \widehat C_{N,N,M}^{+}
    \longrightarrow
    W_2^2(\nu,\widehat\mu_N)
\end{equation}
and
\begin{equation}
    W_2\!\left(
        \widehat\gamma_{N,N,M}^{+},
        \gamma_N^{\rm SDOT}
    \right)
    \longrightarrow0
\end{equation}
as $M\to\infty$.

Since $M_k\to\infty$, the same conclusions hold along the subsequence
$M=M_k$. The event $\Omega_0$ is independent of the particular sequence
$(B_k,M_k)$, so the statement holds simultaneously for every such
sequence.
\end{proof}

\subsection{The nontrivial limit: fixed $B<N$ and $M\to\infty$.}
The more interesting regime keeps the projected-rank restriction fixed while
sending the source resolution to infinity. In this case, ReCouple+ converges
to a population block-constrained semi-discrete OT problem.

Fix a projection direction $\theta\in\mathbb S^{d-1}$ and sort the target
samples according to $\langle\theta,y_i\rangle$, with ties broken
deterministically. Let
\begin{equation}
    I_1,\ldots,I_Q
\end{equation}
denote the resulting consecutive target blocks, with
$b_q:=|I_q|$ and $\sum_{q=1}^Q b_q=N$. Define
\begin{equation}
    \alpha_q:=\frac{b_q}{N},
    \qquad
    s_q:=\sum_{\ell=1}^q\alpha_\ell,
    \qquad
    s_0:=0.
\end{equation}
Let
\begin{equation}
    F_\theta(r)
    :=
    \nu\bigl(
        \{z:\langle\theta,z\rangle\leq r\}
    \bigr)
\end{equation}
be the distribution function of the projected source. Since $\nu$ is
absolutely continuous, its one-dimensional projection is also absolutely
continuous, and in particular $F_\theta$ is continuous and the projected law is atomless.
For $q=1,\ldots,Q-1$, let
\begin{equation}
    c_q:=F_\theta^{-1}(s_q),
\end{equation}
and set $c_0=-\infty$ and $c_Q=+\infty$. Define the population rank slabs
\begin{equation}
\label{eq:population-slabs}
    A_q
    :=
    \left\{
        z:
        c_{q-1}
        <
        \langle\theta,z\rangle
        \leq
        c_q
    \right\}.
\end{equation}
Then
\begin{equation}
    \nu(A_q)=\alpha_q.
\end{equation}
Define the normalized source and target block measures
\begin{equation}
\label{eq:block-measures}
    \nu_q
    :=
    \frac{1}{\alpha_q}\nu|_{A_q},
    \qquad
    \mu_q
    :=
    \frac{1}{b_q}
    \sum_{i\in I_q}\delta_{y_i}.
\end{equation}

We first make precise the corresponding population transport problem.

\begin{proposition}[Population Block-SDOT]
\label{prop:block-sdot}
Let $\Gamma_{B,\theta}$ denote the class of couplings between
$\nu$ and $\widehat\mu_N$ that transport source slab $A_q$ only to
targets belonging to block $I_q$. Then every
$\gamma\in\Gamma_{B,\theta}$ admits the decomposition
\begin{equation}
    \gamma
    =
    \sum_{q=1}^Q
    \alpha_q\gamma_q,
    \qquad
    \gamma_q\in\Pi(\nu_q,\mu_q).
\end{equation}
Consequently,
\begin{equation}
\label{eq:block-sdot-cost}
    \inf_{\gamma\in\Gamma_{B,\theta}}
    \int\|z-y\|^2\,d\gamma(z,y)
    =
    \sum_{q=1}^Q
    \alpha_q
    W_2^2(\nu_q,\mu_q).
\end{equation}
Since each $\nu_q$ is absolutely continuous, the quadratic optimal
coupling $\gamma_q^\star$ between $\nu_q$ and $\mu_q$ is unique. Hence
the constrained problem has the unique optimizer
\begin{equation}
\label{eq:block-sdot}
    \gamma_{B,\theta}^{\rm BSDOT}
    :=
    \sum_{q=1}^Q
    \alpha_q\gamma_q^\star.
\end{equation}
\end{proposition}

\begin{proof}
The source slabs form a partition of $\mathbb R^d$ up to a
$\nu$-null set, and the target blocks form a partition of the target
samples. The block constraint therefore forces exactly $\alpha_q$ units of
mass to be transported from $A_q$ to the targets in $I_q$. Normalizing
the restriction of $\gamma$ to this block gives a coupling
$\gamma_q\in\Pi(\nu_q,\mu_q)$, yielding
\begin{equation}
    \gamma
    =
    \sum_{q=1}^Q\alpha_q\gamma_q.
\end{equation}
The quadratic cost separates accordingly:
\begin{equation}
    \int\|z-y\|^2\,d\gamma
    =
    \sum_{q=1}^Q
    \alpha_q
    \int\|z-y\|^2\,d\gamma_q.
\end{equation}
Minimizing each term independently gives
\eqref{eq:block-sdot-cost}. Since $\nu_q\ll\mathcal L^d$, Brenier's
theorem gives uniqueness of each quadratic optimal coupling
$\gamma_q^\star$, and hence uniqueness of
$\gamma_{B,\theta}^{\rm BSDOT}$.
\end{proof}

The following lemma is the key statistical step.

\begin{lemma}[Convergence of empirical rank blocks]
\label{lem:rank-block-convergence}
Let $(Z_j)_{j\geq1}$ be i.i.d.\ with law $\nu$. For $n=MN$, sort
$Z_1,\ldots,Z_n$ according to
$\langle\theta,Z_j\rangle$. For block $q$, let
$\mathcal I_{q,M}$ contain the source indices having projected ranks
\begin{equation}
    Ms_{q-1}N+1,
    \ldots,
    Ms_qN,
\end{equation}
equivalently the $Mb_q$ consecutive source ranks corresponding to target
block $I_q$, and define
\begin{equation}
\label{eq:empirical-block-measure}
    \widehat\nu_{q,M}
    :=
    \frac{1}{Mb_q}
    \sum_{j\in\mathcal I_{q,M}}\delta_{Z_j}.
\end{equation}
Then, simultaneously for $q=1,\ldots,Q$,
\begin{equation}
\label{eq:block-source-W2}
    W_2(
        \widehat\nu_{q,M},
        \nu_q
    )
    \longrightarrow0
    \qquad
    \text{a.s. as }M\to\infty.
\end{equation}
\end{lemma}

\begin{proof}
Write
\begin{equation}
    R_j:=\langle\theta,Z_j\rangle,
    \qquad
    n:=MN,
\end{equation}
and let
\begin{equation}
    m_q:=\sum_{\ell=1}^q b_\ell,
    \qquad
    r_{q,M}:=Mm_q=ns_q.
\end{equation}
Because the projected law is atomless, with probability one there are no
ties among the $R_j$.

Let
\begin{equation}
    K_{q,M}
    :=
    \sum_{j=1}^n
    \mathbf 1_{\{R_j\leq c_q\}},
\end{equation}
with $K_{0,M}=0$ and $K_{Q,M}=n$. By the strong law of large numbers,
\begin{equation}
    \frac{K_{q,M}}{n}
    \longrightarrow
    F_\theta(c_q)
    =
    s_q
    \qquad
    \text{a.s.}
\end{equation}
Hence
\begin{equation}
\label{eq:rank-count}
    \frac{
        |K_{q,M}-r_{q,M}|
    }{n}
    \longrightarrow0
    \qquad
    \text{a.s.}
\end{equation}

Let
\begin{equation}
    \mathcal J_{q,M}
    :=
    \left\{
        1\leq j\leq n:
        Z_j\in A_q
    \right\}
\end{equation}
be the indices falling in the fixed population slab. The empirical
rank block $\mathcal I_{q,M}$ and the population-slab index set
$\mathcal J_{q,M}$ differ only through discrepancies at their two
boundaries. Therefore
\begin{equation}
\label{eq:sym-difference}
    |\mathcal I_{q,M}\triangle\mathcal J_{q,M}|
    \leq
    |K_{q,M}-r_{q,M}|
    +
    |K_{q-1,M}-r_{q-1,M}|.
\end{equation}
By \eqref{eq:rank-count},
\begin{equation}
    \frac{
        |\mathcal I_{q,M}\triangle\mathcal J_{q,M}|
    }{n}
    \longrightarrow0.
\label{eq:sym-difference-small}
\end{equation}

Let $f$ be bounded and continuous. Since
$Mb_q=n\alpha_q$,
\begin{equation}
\begin{split}
    \int f\,d\widehat\nu_{q,M}
    &=
    \frac{1}{n\alpha_q}
    \sum_{j\in\mathcal I_{q,M}}f(Z_j).
\end{split}
\end{equation}
Using \eqref{eq:sym-difference-small},
\begin{equation}
\begin{split}
    &
    \left|
        \frac{1}{n\alpha_q}
        \sum_{j\in\mathcal I_{q,M}}f(Z_j)
        -
        \frac{1}{n\alpha_q}
        \sum_{j=1}^n
        f(Z_j)\mathbf 1_{\{Z_j\in A_q\}}
    \right| \\
    &\qquad\leq
    \frac{\|f\|_\infty}{
        n\alpha_q
    }
    |\mathcal I_{q,M}\triangle\mathcal J_{q,M}|
    \longrightarrow0.
\end{split}
\end{equation}
The strong law gives
\begin{equation}
    \frac{1}{n\alpha_q}
    \sum_{j=1}^n
    f(Z_j)\mathbf 1_{\{Z_j\in A_q\}}
    \longrightarrow
    \frac{
        \mathbb E[
            f(Z)\mathbf 1_{\{Z\in A_q\}}
        ]
    }{
        \alpha_q
    }
    =
    \int f\,d\nu_q.
\end{equation}
Hence
\begin{equation}
    \widehat\nu_{q,M}
    \rightharpoonup
    \nu_q.
\label{eq:block-weak}
\end{equation}

It remains to prove convergence of second moments. Let
\begin{equation}
    H_j:=\|Z_j\|^2.
\end{equation}
The difference between the second moment over the empirical rank block and
that over the fixed slab is bounded by
\begin{equation}
\label{eq:moment-difference}
    \frac{1}{n\alpha_q}
    \sum_{
        j\in
        \mathcal I_{q,M}\triangle\mathcal J_{q,M}
    }
    H_j.
\end{equation}
For any $K>0$,
\begin{equation}
\begin{split}
    \frac1n
    \sum_{
        j\in
        \mathcal I_{q,M}\triangle\mathcal J_{q,M}
    }
    H_j
    &\leq
    K\,
    \frac{
        |\mathcal I_{q,M}\triangle\mathcal J_{q,M}|
    }{n} +
    \frac1n
    \sum_{j=1}^n
    H_j\mathbf 1_{\{H_j>K\}}.
\end{split}
\end{equation}
The first term converges to zero by
\eqref{eq:sym-difference-small}. By the strong law, the limsup of the
second term equals
\begin{equation}
    \mathbb E[
        \|Z\|^2
        \mathbf 1_{\{\|Z\|^2>K\}}
    ].
\end{equation}
Since $\nu\in\mathcal P_2(\mathbb R^d)$, this quantity tends to zero as
$K\to\infty$. Therefore \eqref{eq:moment-difference} converges to zero.

A final application of the strong law of large number gives
\begin{equation}
    \frac{1}{n\alpha_q}
    \sum_{j=1}^n
    \|Z_j\|^2
    \mathbf 1_{\{Z_j\in A_q\}}
    \longrightarrow
    \int\|z\|^2\,d\nu_q(z).
\end{equation}
Thus
\begin{equation}
    \int\|z\|^2\,d\widehat\nu_{q,M}(z)
    \longrightarrow
    \int\|z\|^2\,d\nu_q(z).
\end{equation}
Together with \eqref{eq:block-weak}, this is equivalent to
\eqref{eq:block-source-W2}.
\end{proof}

We can now state the fixed-block limit.

\begin{theorem}[Fixed-block ReCouple+ converges to Block-SDOT]
\label{thm:fixed-B-block-sdot}
Let $\nu\in\mathcal P_2(\mathbb R^d)$ be absolutely continuous, and fix
$\widehat\mu_N$, $\theta$, and the target blocks
$I_1,\ldots,I_Q$ defined above. Construct ReCouple+ from the nested source
sequence $(Z_j)_{j\geq1}$ by sorting the $MN$ pooled source samples
according to $\langle\theta,Z_j\rangle$, assigning the $Mb_q$
corresponding consecutive source ranks to target block $I_q$, and solving
the capacitated assignment problem within each block.

Let $\widehat\gamma_{N,B,M}^{+,\theta}$ denote any resulting optimal
block-constrained coupling, and let
$\widehat C_{N,B,M}^{+,\theta}$ denote its quadratic cost. Then there
exists an event $\Omega_0$ with $\mathbb P(\Omega_0)=1$ such that, on
$\Omega_0$, for every choice of finite-$M$ blockwise optimal couplings,
\begin{equation}
\label{eq:block-plan-limit}
    W_2
    \left(
        \widehat\gamma_{N,B,M}^{+,\theta},
        \gamma_{B,\theta}^{\rm BSDOT}
    \right)
    \longrightarrow0
    \qquad
    \text{as }M\to\infty,
\end{equation}
and
\begin{equation}
\label{eq:block-cost-limit}
    \widehat C_{N,B,M}^{+,\theta}
    \longrightarrow
    \sum_{q=1}^Q
    \alpha_q
    W_2^2(\nu_q,\mu_q).
\end{equation}
\end{theorem}

\begin{proof}
Fix an outcome in the probability-one event on which
Lemma~\ref{lem:rank-block-convergence} holds simultaneously for all
$q=1,\ldots,Q$. We argue deterministically on this event.

\medskip
\noindent
\textbf{Step 1: Each finite block problem is exact empirical OT.}

For block $q$, the empirical source measure
$\widehat\nu_{q,M}$ contains $Mb_q$ equally weighted source samples.
Each target $y_i$, $i\in I_q$, has capacity $M$. Replicating every target
in $I_q$ exactly $M$ times converts the capacitated problem into an
equal-weight assignment between $Mb_q$ source samples and $Mb_q$ target
slots.

By the Birkhoff--von Neumann theorem, the minimum of the corresponding
transport linear program is attained at a permutation. Therefore the
blockwise assignment solved by ReCouple+ is exactly the quadratic optimal
transport problem
\begin{equation}
    \widehat\gamma_{q,M}
    \in
    \Pi_{\rm opt}
    (
        \widehat\nu_{q,M},
        \mu_q
    ),
\end{equation}
and its normalized block cost is
\begin{equation}
\label{eq:block-emp-cost}
    C_{q,M}
    =
    W_2^2(
        \widehat\nu_{q,M},
        \mu_q
    ).
\end{equation}

The global empirical coupling decomposes as
\begin{equation}
\label{eq:empirical-global-decomp}
    \widehat\gamma_{N,B,M}^{+,\theta}
    =
    \sum_{q=1}^Q
    \alpha_q
    \widehat\gamma_{q,M},
\end{equation}
and therefore
\begin{equation}
\label{eq:empirical-global-cost}
    \widehat C_{N,B,M}^{+,\theta}
    =
    \sum_{q=1}^Q
    \alpha_q C_{q,M}.
\end{equation}

\medskip
\noindent
\textbf{Step 2: Convergence of each block cost.}

By Lemma~\ref{lem:rank-block-convergence},
\begin{equation}
    W_2(
        \widehat\nu_{q,M},
        \nu_q
    )
    \longrightarrow0.
\end{equation}
Since $\mu_q$ is fixed, the triangle inequality gives
\begin{equation}
    \left|
        W_2(
            \widehat\nu_{q,M},
            \mu_q
        )
        -
        W_2(
            \nu_q,
            \mu_q
        )
    \right|
    \leq
    W_2(
        \widehat\nu_{q,M},
        \nu_q
    )
    \longrightarrow0.
\end{equation}
Hence
\begin{equation}
    C_{q,M}
    \longrightarrow
    W_2^2(
        \nu_q,
        \mu_q
    ).
\label{eq:block-cost-convergence}
\end{equation}
Using \eqref{eq:empirical-global-cost} and the fact that $Q$ is finite,
\begin{equation}
    \widehat C_{N,B,M}^{+,\theta}
    \longrightarrow
    \sum_{q=1}^Q
    \alpha_q
    W_2^2(
        \nu_q,
        \mu_q
    ),
\end{equation}
which proves \eqref{eq:block-cost-limit}.

\medskip
\noindent
\textbf{Step 3: Convergence of each block coupling.}

Fix $q$. The marginals of $\widehat\gamma_{q,M}$ are
$\widehat\nu_{q,M}$ and $\mu_q$, whose second moments are uniformly
bounded. Hence $\{\widehat\gamma_{q,M}\}_M$ is tight.

Consider an arbitrary subsequence. By Prokhorov's theorem it admits a
further weakly convergent subsequence
\begin{equation}
    \widehat\gamma_{q,M_\ell}
    \rightharpoonup
    \bar\gamma_q.
\end{equation}
The convergence of the marginals implies
\begin{equation}
    \bar\gamma_q
    \in
    \Pi(
        \nu_q,
        \mu_q
    ).
\end{equation}
By lower semicontinuity of the quadratic cost and
\eqref{eq:block-cost-convergence},
\begin{equation}
\begin{split}
    \int\|z-y\|^2\,d\bar\gamma_q
    &\leq
    \liminf_{\ell\to\infty}
    \int\|z-y\|^2
    \,d\widehat\gamma_{q,M_\ell}\\
    &=
    W_2^2(
        \nu_q,
        \mu_q
    ).
\end{split}
\end{equation}
The reverse inequality follows because
$\bar\gamma_q\in\Pi(\nu_q,\mu_q)$. Hence $\bar\gamma_q$ is optimal.

Since $\nu_q$ is absolutely continuous, the quadratic optimal coupling
between $\nu_q$ and $\mu_q$ is unique. Therefore
\begin{equation}
    \bar\gamma_q
    =
    \gamma_q^\star.
\end{equation}
Every subsequence therefore admits a further subsequence converging to
$\gamma_q^\star$, and consequently
\begin{equation}
    \widehat\gamma_{q,M}
    \rightharpoonup
    \gamma_q^\star.
\end{equation}

Moreover,
\begin{equation}
\begin{split}
    \int
    (
        \|z\|^2+\|y\|^2
    )
    \,d\widehat\gamma_{q,M}
    &=
    \int\|z\|^2
    \,d\widehat\nu_{q,M}
    +
    \int\|y\|^2
    \,d\mu_q\\
    &\longrightarrow
    \int\|z\|^2\,d\nu_q
    +
    \int\|y\|^2\,d\mu_q.
\end{split}
\end{equation}
Thus the weak convergence upgrades to
\begin{equation}
\label{eq:block-plan-W2}
    W_2(
        \widehat\gamma_{q,M},
        \gamma_q^\star
    )
    \longrightarrow0
\end{equation}
on $\mathbb R^d\times\mathbb R^d$.

\medskip
\noindent
\textbf{Step 4: Reassemble the blocks.}

Using
\eqref{eq:empirical-global-decomp} and
\eqref{eq:block-sdot},
\begin{equation}
    \widehat\gamma_{N,B,M}^{+,\theta}
    =
    \sum_{q=1}^Q
    \alpha_q\widehat\gamma_{q,M},
    \qquad
    \gamma_{B,\theta}^{\rm BSDOT}
    =
    \sum_{q=1}^Q
    \alpha_q\gamma_q^\star.
\end{equation}
For each $q$, let $\Lambda_{q,M}$ be an optimal $W_2$ coupling between
$\widehat\gamma_{q,M}$ and $\gamma_q^\star$. Then
\begin{equation}
    \Lambda_M
    :=
    \sum_{q=1}^Q
    \alpha_q\Lambda_{q,M}
\end{equation}
is a coupling between
$\widehat\gamma_{N,B,M}^{+,\theta}$ and
$\gamma_{B,\theta}^{\rm BSDOT}$. Therefore
\begin{equation}
\begin{split}
    W_2^2
    \left(
        \widehat\gamma_{N,B,M}^{+,\theta},
        \gamma_{B,\theta}^{\rm BSDOT}
    \right)
    &\leq
    \sum_{q=1}^Q
    \alpha_q
    W_2^2(
        \widehat\gamma_{q,M},
        \gamma_q^\star
    ) \longrightarrow0,
\end{split}
\end{equation}
using \eqref{eq:block-plan-W2} and the finiteness of $Q$.
This proves \eqref{eq:block-plan-limit}.
\end{proof}

\begin{corollary}[Block-SDOT cost gap]
\label{cor:block-sdot-gap}
The Block-SDOT cost satisfies
\begin{equation}
\label{eq:block-sdot-gap}
    \sum_{q=1}^Q
    \alpha_q
    W_2^2(
        \nu_q,\mu_q
    )
    \geq
    W_2^2(
        \nu,\widehat\mu_N
    ).
\end{equation}
Equality holds if and only if the unique SDOT coupling
$\gamma_N^{\rm SDOT}$ respects the block constraint, in which case
\begin{equation}
    \gamma_{B,\theta}^{\rm BSDOT}
    =
    \gamma_N^{\rm SDOT}.
\end{equation}
\end{corollary}

\begin{proof}
The admissible Block-SDOT class
$\Gamma_{B,\theta}$ is a subset of the full coupling class
$\Pi(\nu,\widehat\mu_N)$, so restricting the feasible set can only
increase the optimal quadratic cost. This proves
\eqref{eq:block-sdot-gap}. If equality holds, the unique optimizer of the
unrestricted problem is also feasible for the constrained problem, hence
must equal $\gamma_{B,\theta}^{\rm BSDOT}$. The converse is immediate.
\end{proof}

\subsection{Spatial Regularity for SDOT}

\begin{theorem}[Spatial regularity for SDOT]
\label{prop:app_straight-not-spatially-simple}
Let $T=T_N^\star$ be the SDOT map from an absolutely
continuous source $\nu$ to
$\widehat\mu_N=\frac1N\sum_{i=1}^N\delta_{y_i}$,
and let $\{L_i\}_{i=1}^N$ denote its Laguerre cells, so that
$T(z)=y_i$ for $\nu$-a.e.\ $z\in L_i$.
Consider the displacement interpolation
$X_t=(1-t)Z+tT(Z), \  0\le t<1,$
and let $v_t^\star$ denote its marginal velocity field. For each $i$,
define 
$L_i^t:=(1-t)L_i+t y_i.$

\begin{itemize}[leftmargin=18pt, nosep]
\item[(i)] {\it Cellwise regularity.}
For $t\in[0,1)$, the marginal velocity admits the representation $\bar v_t^\star(x) = \frac{y_i-x}{1-t}$ for $x\in\operatorname{int}(L_i^t),$ which agrees with $v_t^\star$ $p_t$-a.e.,
and hence $\nabla_x\bar v_t^\star(x)
=
-\frac{1}{1-t}I_d$,
and 
\begin{equation}
\|\nabla_x\bar v_t^\star(x)\|_{\mathrm{op}}
=
\frac{1}{1-t},
\qquad
\|\nabla_x\bar v_t^\star(x)\|_F^2
=
\frac{d}{(1-t)^2}.
\label{eq:sdot-cellwise-jacobian-norm}
\end{equation}

\item[(ii)] {\it Global stitching barrier.}
Suppose that two distinct Laguerre cells $L_i$ and $L_j$, with
$y_i\neq y_j$, have a common boundary point,
$\overline{L_i}\cap\overline{L_j}\neq\emptyset.$
Fix $t\in(0,1)$, and let
$\widetilde v_t:\mathbb R^d\to\mathbb R^d$ be any globally Lipschitz
field agreeing with  $\bar v_t^\star$ on the
interiors of the  cells. Then
$\operatorname{Lip}(\widetilde v_t)
\ge
\max\left\{
\frac1t,\frac1{1-t}
\right\}.$
In particular, if $\widetilde v_t\in C^1(\mathbb R^d;\mathbb R^d)$,
then
\begin{equation}
\sup_{x\in\mathbb R^d}
\|\nabla_x\widetilde v_t(x)\|_{\mathrm{op}}
\ge
\max\left\{
\frac1t,\frac1{1-t}
\right\}, \quad
\sup_{x\in\mathbb R^d}
\|\nabla_x\widetilde v_t(x)\|_F
\ge
\max\left\{
\frac1t,\frac{\sqrt d}{1-t}
\right\}.
\label{eq:sdot-global-frobenius-lower}
\end{equation}
\end{itemize}
\end{theorem}

\begin{proof}
For part~(i), let $z\in\operatorname{int}(L_i)$. Since $T(z)=y_i$,
\begin{equation}
x=(1-t)z+t y_i.
\end{equation}
For $t<1$, this can be rewritten as:
\begin{equation}
z=\frac{x-t y_i}{1-t}.
\end{equation}
The velocity along the corresponding displacement trajectory is
$T(z)-z$, and hence
\begin{equation}
T(z)-z
=
y_i-\frac{x-t y_i}{1-t}
=
\frac{y_i-x}{1-t}.
\end{equation}

Since the OT map for quadratic cost is monotone, the interpolation map
$S_t=(1-t)I+tT$ is strongly monotone for $t<1$:
\begin{equation}
\langle S_t(z)-S_t(z'),z-z'\rangle
\ge
(1-t)\|z-z'\|^2.
\end{equation}
Thus $S_t$ is injective, up to the usual null-set ambiguity of $T$,
so the transported cell interiors define unambiguous branches of the
marginal velocity. Differentiating
gives 
\eqref{eq:sdot-cellwise-jacobian-norm}.

For part~(ii), the within-cell identity already implies
\begin{equation}
\operatorname{Lip}(\widetilde v_t)
\ge
\frac1{1-t}.
\end{equation}
Indeed, for any $x,x'\in\operatorname{int}(L_i^t)$,
\begin{equation}
\|\bar v_t^\star(x)-\bar v_t^\star(x')\|
=
\frac{1}{1-t}\|x-x'\|.
\end{equation}

It remains to establish the cross-cell bound. Choose
$z_\star\in\overline{L_i}\cap\overline{L_j}$.
Approaching $z_\star$ from the two cell interiors produces the
corresponding limiting points
\begin{equation}
x_i^\star=(1-t)z_\star+t y_i,
\qquad
x_j^\star=(1-t)z_\star+t y_j.
\end{equation}
Their spatial separation is
\begin{equation}
\|x_i^\star-x_j^\star\|
=
t\|y_i-y_j\|.
\label{eq:sdot-boundary-position-gap}
\end{equation}
The limiting cellwise velocities are
\begin{equation}
v_i^\star=y_i-z_\star,
\qquad
v_j^\star=y_j-z_\star,
\end{equation}
and hence
\begin{equation}
\|v_i^\star-v_j^\star\|
=
\|y_i-y_j\|.
\label{eq:sdot-boundary-velocity-gap}
\end{equation}
Since $\widetilde v_t$ is Lipschitz, it is continuous; because it
agrees with $\bar v_t^\star$ on the cell interiors, it therefore takes
these limiting values at $x_i^\star$ and $x_j^\star$. Consequently,
\begin{equation}
\operatorname{Lip}(\widetilde v_t)
\ge
\frac{
\|\widetilde v_t(x_i^\star)-\widetilde v_t(x_j^\star)\|
}{
\|x_i^\star-x_j^\star\|
}
=
\frac1t.
\end{equation}
Together with the within-cell bound, this proves the first lower bound. 

If $\widetilde v_t$ is $C^1$, then on $\mathbb R^d$
\begin{equation}
\operatorname{Lip}(\widetilde v_t)
=
\sup_x\|\nabla_x\widetilde v_t(x)\|_{\mathrm{op}}
\end{equation}
whenever the right-hand side is finite, which gives  the second lower bound.

Moreover,
$\|A\|_F\ge\|A\|_{\mathrm{op}}$, while part~(i) gives the exact
within-cell value
\begin{equation}
\|\nabla_x\bar v_t^\star(x)\|_F
=
\frac{\sqrt d}{1-t}.
\end{equation}
This proves \eqref{eq:sdot-global-frobenius-lower}.

Finally, we remark that at $t=0$ the two limiting spatial points coincide,
$x_i^\star=x_j^\star=z_\star$, whereas their limiting velocities differ
by
\begin{equation}
v_i^\star-v_j^\star
=
y_i-y_j\neq0.
\end{equation}
Hence no continuous field can agree with both cellwise branches at the
common boundary.
\end{proof}

\subsection{Proof of Proposition \ref{prop:recouple-randomblock}}
\label{app_recouple-randomblock}

\begin{proof} We prove each statement separately. 

\paragraph{Proof of (i)} There are 
$K=\frac{N}{B}$ matched source--target block pairs. Each pair $S_k\times T_k$ contains exactly $B^2$ admissible edges. Since the block pairs are disjoint, \begin{equation} |E_{\mathcal B}| = \sum_{k=1}^K |S_k\times T_k| = KB^2 = NB. \end{equation} Dividing by the $N^2$ edges of the complete bipartite graph gives edge density \begin{equation} \frac{|E_{\mathcal B}|}{N^2} = \frac{B}{N}. \end{equation} 

\paragraph{Proof of (ii)} Condition on the source partition $S_1,\ldots,S_K$. Fix a source index $i$, and let $S_k$ be the unique source block containing $i$. The target block $T_k$ paired with $S_k$ is a uniformly random subset of $B$ target indices. Therefore, for the fixed target $\pi^\star(i)$, \begin{equation} \mathbb P \left( \pi^\star(i)\in T_k \right) = \frac{B}{N}. \end{equation} Equivalently, \begin{equation} \mathbb P \left( (i,\pi^\star(i)) \in E_{\mathcal B} \right) = \frac{B}{N}. \end{equation} By linearity of expectation, \begin{align} \mathbb E[ \operatorname{Rec}^{\rm rand}_B ] &= \frac1N \sum_{i=1}^N \mathbb P \left( (i,\pi^\star(i)) \in E_{\mathcal B} \right) \nonumber\\ &= \frac1N \sum_{i=1}^N \frac{B}{N} = \frac{B}{N}, \end{align} 
which gives ~\eqref{eq:random-recall-prop}. Now suppose that $\pi^\star$ is the unique global OT permutation. Then \begin{equation} C_B^{\rm rand}=C^\star \end{equation} if and only if $\pi^\star$ remains feasible under the random block restriction. Conditional on the source partition, this requires \begin{equation} T_k=\pi^\star(S_k), \qquad k=1,\ldots,K. \end{equation} Thus exactly one labeled target partition is compatible with $\pi^\star$. The number of balanced labeled target partitions is \begin{equation} \binom{N}{B} \binom{N-B}{B} \cdots \binom{B}{B} = \frac{N!}{(B!)^K}. \end{equation} Since RandomBlock draws uniformly from these partitions, \begin{equation} \mathbb P( C_B^{\rm rand}=C^\star ) = \frac{1}{ N!/(B!)^K } = \frac{(B!)^K}{N!}. \end{equation} Using $K=N/B$ yields~\eqref{eq:random-exact-prop}.

\paragraph{Proof of (iii)} Set \begin{equation} a_r:=\theta^\top x_{i_r}, \qquad b_r:=\theta^\top y_{j_r}, \end{equation} so that \begin{equation} a_1<\cdots<a_N, \qquad b_1<\cdots<b_N. \end{equation} We first show that matching $a_r$ to $b_r$ is optimal for the one-dimensional squared cost. Consider any assignment containing an inversion: for some $r<s$ it matches \begin{equation} a_r\longleftrightarrow b_\ell, \qquad a_s\longleftrightarrow b_k, \qquad k<\ell. \end{equation} The difference between the cost before and after uncrossing these two matches is \begin{align} &\left[ (a_r-b_\ell)^2 + (a_s-b_k)^2 \right] - \left[ (a_r-b_k)^2 + (a_s-b_\ell)^2 \right] \nonumber\\ &\qquad= 2(a_s-a_r)(b_\ell-b_k) \geq0. \end{align} Thus removing an inversion never increases the projected quadratic cost. Repeatedly uncrossing inversions yields the monotone rank matching \begin{equation} a_r\longleftrightarrow b_r, \qquad r=1,\ldots,N, \end{equation} which is therefore an optimal one-dimensional quadratic coupling. Now partition the two rank orderings into corresponding consecutive blocks of size $B$. The permutation $\sigma_\theta$ matches each rank $r$ to the same rank $r$ and is therefore feasible for the ReCouple block geometry. 
Hence 
\begin{equation} C_B^\theta \leq C_{\rm rank}^\theta. \end{equation} On the other hand, the ReCouple feasible set is a subset of all permutations, \begin{equation} \mathcal F_B^\theta \subseteq S_N,
\end{equation} 
so restricting the feasible set cannot improve upon unrestricted OT: \begin{equation} C^\star \leq C_B^\theta. \end{equation} Together these give~\eqref{eq:rank-sandwich-prop}. If $B=1$, every block contains a single source rank and a single target rank, so $\sigma_\theta$ is the only feasible permutation and \begin{equation} C_1^\theta = C_{\rm rank}^\theta. \end{equation} If $B=N$, there is one global block, every permutation is feasible, and \begin{equation} C_N^\theta = C^\star. \end{equation} 

\paragraph{Proof of (iv)} The dual of the unrestricted assignment problem is \begin{equation} \max_{u,v} \frac1N \left( \sum_{i=1}^N u_i + \sum_{j=1}^N v_j \right) \end{equation} subject to \begin{equation} u_i+v_j\leq c_{ij}. \end{equation} Let $(u^\star,v^\star)$ be any optimal dual pair and define $r_{ij}$ by~\eqref{eq:reduced-cost-prop}. Dual feasibility gives $r_{ij}\geq0$. 
For any permutation $\pi\in S_N$, \begin{align} \frac1N \sum_{i=1}^N c_{i,\pi(i)} &= \frac1N \sum_{i=1}^N \left( u_i^\star + v_{\pi(i)}^\star + r_{i,\pi(i)} \right) \nonumber\\ &= \frac1N \left( \sum_{i=1}^N u_i^\star + \sum_{j=1}^N v_j^\star \right) + \frac1N \sum_{i=1}^N r_{i,\pi(i)}, \end{align} where the second equality uses the fact that $\pi$ is a permutation, so \begin{equation} \sum_{i=1}^N v_{\pi(i)}^\star = \sum_{j=1}^N v_j^\star. \end{equation} By strong duality, \begin{equation} \frac1N \left( \sum_i u_i^\star + \sum_j v_j^\star \right) = C^\star. \end{equation} 

Therefore, for every permutation, \begin{equation} \frac1N \sum_i c_{i,\pi(i)} - C^\star = \frac1N \sum_i r_{i,\pi(i)}. \end{equation} Minimizing this identity over $\pi\in\mathcal F_{\mathcal B}$ gives \begin{equation} C_{\mathcal B}-C^\star = \frac1N \min_{\pi\in\mathcal F_{\mathcal B}} \sum_i r_{i,\pi(i)}, \end{equation} which proves~\eqref{eq:restricted-reduced-cost}. 
\end{proof}

\section{Details on the Empirical Results and Ablation Studies}
\label{app_experimental}

\subsection{2D Multimodal Task} \label{app_2ddata}

\paragraph{Data and pretrained model}
We use a two-dimensional target consisting of $25$ isotropic Gaussian
modes arranged on a $5\times5$ grid, with spacing $0.5$ and standard
deviation $0.05$, rotated by $45^\circ$ and normalized following the
dataset generator.  We use $50{,}000$ training samples and an
independent $50{,}000$-sample test set, with Gaussian source
$\nu=\mathcal N(0,I_2)$.  The velocity field is a Fourier-feature MLP
with four hidden layers of width $256$ and SiLU activations, using
$64$-dimensional spatial and time embeddings.  The base model is
pretrained with independent CFM, and all post-training methods start
from the same validated EMA checkpoint.

\paragraph{ReCouple family}
ReCouple uses a persistent Gaussian source bank, block size $B=512$,
and $R=16$ random projection proposals.  For each proposal, source and
target samples are sorted by their projected coordinates, divided into
corresponding blocks, and matched by the assignment in the
original data space.  The candidate with the smallest full-bank
quadratic cost is retained.  We vary
$B\in\{512,1024,2048\}$ in the block-size ablation below.
ReCouple+ uses the same projected geometry with source multiplicity
$M=2$, assigning $MN$ Gaussian source samples to $N$ targets with
capacity $M$ per target.  The projection selected at $M=1$ is retained
to isolate the effect of source resolution.

ReCouple++ uses the same pooled coupling as ReCouple+ and adds the
spatial consistency penalty
\[
\mathcal R_\sigma
=
\frac{1}{2\sigma^2}
\mathbb E_\epsilon
\left\|
v_\theta(x_t+\sigma\epsilon,t)-v_\theta(x_t,t)
\right\|^2,
\qquad
\epsilon\sim\mathcal N(0,I_2).
\]
We set $\sigma$ to $5\%$ of the data scale, calibrate its initial
contribution to $25\%$ of the CFM loss, and cosine-anneal its weight to
zero over the first $70\%$ of post-training.

\paragraph{Baselines}
We compare against the following matched baselines. In Figure \ref{fig:main_visual}, \emph{Pretrained}
denotes the independently pretrained CFM model without post-training.
\emph{PersistentIndependent-FT} continues training on one fixed
independent Gaussian--target pairing, thereby controlling for additional
optimization on a persistent pair bank. \emph{OT-CFM-FT}
\citep{tong2023improving} instead constructs fresh exact minibatch OT
couplings at every post-training update. 
We also evaluate \emph{AlignFlow}
\citep{kong2026alignflow}, which uses its coupling-alignment construction with
the same pretrained backbone and post-training budget. \emph{SDOT}
\citep{mousavi2026flow} uses the global semi-discrete quadratic OT
coupling from the continuous Gaussian source to the empirical target,
providing the unrestricted OT reference. Finally,
\emph{Noise-ReCouple} (cf.\ \citealp{hui2025not}) keeps the selected
ReCouple pairing but perturbs its source endpoints according to
\[
z_\beta=\sqrt{1-\beta}\,z+\sqrt{\beta}\,\epsilon,
\qquad
\epsilon\sim\mathcal N(0,I_2),
\]
with $\beta=0.01$, providing a control for improving learnability by
source smoothing rather than spatial regularization. Whenever
applicable, baselines use the same network architecture, pretrained
initialization, optimizer, update budget, EMA, and evaluation protocol.

\paragraph{Baseline implementation details} OT-CFM-FT uses fresh exact minibatch OT couplings of size $128$ at every update. Noise-ReCouple uses the same persistent ReCouple pairs with source perturbation parameter $\beta=0.01$. AlignFlow follows the published 2D coupling construction, using an entropic dual fit and a balanced fixed Gaussian pair bank, followed by the common post-training protocol. For SDOT, we solve the unrestricted semi-discrete quadratic OT problem from $\mathcal N(0,I_2)$ to the full empirical target using OTT-JAX \citep{cuturi2022optimal}; during  post-training, fresh Gaussian samples are mapped through the resulting fixed Laguerre map at every update. Thus the reported SDOT results correspond to the post-trained neural network rather than direct evaluation of the SDOT map. 

Given a pretrained FM backbone, teacher Reflow and distillation method use 500k teacher-generated pairs
and 50k student updates. Distillation uses endpoint MSE without
LPIPS or a preceding reflow stage. 
As for Reflow,  note that the training can be improved further by using a stronger backbone, adding  task-specific training tricks and increase compute budgets (see, e.g.,  \citep{lee2024improving}) to narrow the gap to state-of-the-art distillation schemes. However, doing so would introduce several additional confounding factors, making it much harder to isolate the effect of the mechanism studied here.

\paragraph{Post-training and evaluation}
All fine-tuned models use $10{,}000$ updates, batch size $2048$, AdamW
with learning rate $10^{-4}$ and weight decay $10^{-5}$, gradient
clipping at $10$, and EMA decay $0.999$.  Generation uses fresh
Gaussian initial conditions and uniform Euler integration at
$\mathrm{NFE}\in\{1,2,4,8\}$.  We generate $20{,}000$ samples per
method.  Empirical $W_2$ is computed by exact Hungarian matching on
$1000$-point real/generated subsamples and averaged over three fixed
subsampling seeds.

We additionally report the neural material-acceleration energy
\[
\mathcal E_{\mathrm{acc}}
=
\int_0^1
\mathbb E
\left\|
\partial_t v_t(X_t)
+
\nabla_xv_t(X_t)v_t(X_t)
\right\|^2dt,
\]
estimated by automatic differentiation along $512$ common Gaussian
trajectories using $100$ uniform Euler steps.

\paragraph{Coupling resolution versus SDOT}
Table~\ref{tab:2d-coupling-construction} compares the endpoint
transport objective and construction cost of pooled ReCouple with
global SDOT.  SDOT attains a slightly smaller quadratic transport cost in this experiment,
whereas ReCouple+  is substantially cheaper to
construct.  This comparison illustrates the intended role of
$(B,M)$: ReCouple+ provides a controlled intermediate approximation
rather than an attempt to outperform a dedicated SDOT solver in
transport cost.

\begin{table}[H]
\centering
\caption{
Coupling-construction diagnostics on the 2D task. Preparation times 
exclude neural post-training.
}
\label{tab:2d-coupling-construction}
\vspace{0.2cm}
\small
\setlength{\tabcolsep}{5pt}
\begin{tabular}{lrrr}
\toprule
Coupling
& Quadratic cost $\downarrow$
& Source/target representation
& Prep. time (s) $\downarrow$ \\
\midrule
ReCouple+ ($B=512,M=2$)
& 0.7234
& $100$k source / $50$k target
& 159.55 \\
Global SDOT
& \textbf{0.6692}
& continuous source / $50$k target
& 815.19 \\
\bottomrule
\end{tabular}
\end{table}

\paragraph{Additional distributional diagnostics}
Tables~\ref{tab:2d-metrics-nfe1}--\ref{tab:2d-metrics-nfe8} show that
the low $W_2$ values of the ReCouple family are accompanied by strong
mode coverage rather than concentration on a subset of modes.  At
NFE~1, ReCouple and its variants substantially improve recall and
coverage over the pretrained and persistent-independent models, with
ReCouple achieving coverage $0.7357$ and ReCouple+ the smallest sliced
$W_2$.  By NFE~8, all ReCouple variants attain recall above $0.99$ and
coverage around $0.90$.

The comparison with SDOT is also informative.  SDOT attains the
strongest precision/recall-based scores at NFE~8, while the ReCouple
family obtains substantially smaller sliced $W_2$.  Thus improved
endpoint transport cost or trajectory straightness does not induce a
uniform ordering across distributional metrics after neural
realization.  This is consistent with our motivation for treating
coupling resolution and spatial realizability as distinct design
considerations.

\begin{table}[H]
\centering
\caption{
Additional distributional metrics on the 2D task at NFE~1.
Higher is better for precision, recall, density, and coverage;
lower is better for sliced Wasserstein distance (sW2).
}
\label{tab:2d-metrics-nfe1}
\vspace{0.2cm}
\small
\setlength{\tabcolsep}{4.5pt}
\begin{tabular}{lrrrrr}
\toprule
Method
& Precision $\uparrow$
& Recall $\uparrow$
& Density $\uparrow$
& Coverage $\uparrow$
& sW2 $\downarrow$ \\
\midrule
Pretrained
& {\bf 0.9934} & 0.1138 & {\bf 0.9979} & 0.0423 & 0.4293 \\
PersistentIndependent-FT
& 0.9824 & 0.2443 & 0.9843 & 0.0410 & 0.4163 \\
AlignFlow
& 0.5872 & 0.1893 & 0.4590 & 0.0449 & 0.3654 \\
OT-CFM-FT
& 0.5789 & 0.8830 & 0.3972 & 0.5273 & 0.0384 \\
SDOT
& 0.8700 & \textbf{0.9559} & 0.7973 & 0.7264 & 0.0373 \\
\midrule
Noise-ReCouple
& 0.7038 & 0.9459 & 0.5610 & 0.6626 & 0.0292 \\
ReCouple
& 0.8107 & 0.9477 & 0.7124 & \textbf{0.7357} & 0.0236 \\
ReCouple-AS
& 0.8077 & 0.9498 & 0.7021 & 0.7221 & 0.0236 \\
ReCouple+
& 0.8191 & 0.9467 & 0.7196 & 0.7318 & \textbf{0.0233} \\
ReCouple++
& 0.8052 & 0.9493 & 0.7019 & 0.7246 & 0.0235 \\
\bottomrule
\end{tabular}
\end{table}

\begin{table}[H]
\centering
\caption{
Additional distributional metrics on the 2D task at NFE~8.
Higher is better for precision, recall, density, and coverage;
lower is better for sliced Wasserstein distance (sW2).
}
\label{tab:2d-metrics-nfe8}
\vspace{0.2cm}
\small
\setlength{\tabcolsep}{4.5pt}
\begin{tabular}{lrrrrr}
\toprule
Method
& Precision $\uparrow$
& Recall $\uparrow$
& Density $\uparrow$
& Coverage $\uparrow$
& sW2 $\downarrow$ \\
\midrule
Pretrained
& 0.5622 & 0.8803 & 0.3928 & 0.5298 & 0.0732 \\
PersistentIndependent-FT
& 0.5492 & 0.8315 & 0.3796 & 0.5073 & 0.0748 \\
AlignFlow
& 0.6503 & 0.9343 & 0.5044 & 0.6226 & 0.0609 \\
OT-CFM-FT
& 0.9125 & 0.9781 & 0.8713 & 0.8437 & 0.0176 \\
SDOT
& \textbf{0.9332} & \textbf{0.9948} & \textbf{0.8901}
& \textbf{0.9137} & 0.0381 \\
\midrule
Noise-ReCouple
& 0.9131 & 0.9920 & 0.8650 & 0.9013 & {\bf 0.0171} \\
ReCouple
& 0.9110 & 0.9912 & 0.8677 & 0.8999 & 0.0172 \\
ReCouple-AS
& 0.9072 & 0.9945 & 0.8589 & 0.9016 & 0.0173 \\
ReCouple+
& 0.9122 & 0.9917 & 0.8700 & 0.9058 & 0.0173 \\
ReCouple++
& 0.9056 & 0.9915 & 0.8576 & 0.9046 & 0.0173 \\
\bottomrule
\end{tabular}
\end{table}

\subsection{CIFAR-10 Task} \label{app_cifar10ablate}

\paragraph{Data and pretrained model}
We consider unconditional generation of CIFAR-10 images at $32\times32$ resolution, using the 50,000 training images normalized to $[-1,1]$. We use two pretrained backbones: rectified flow (RF), trained with independent couplings, and OT-CFM, trained with exact minibatch OT. Both use the same U-Net with 128 base channels, channel multipliers $(1,2,2,2)$, four residual blocks per resolution, and dropout $0.15$, pretrained for 234,375 updates with seed 10. Post-training starts from the corresponding EMA weights. ReCouple uses the fixed, unflipped training set and standard Gaussian source samples.

\paragraph{ReCouple family}
The default ReCouple configuration uses block size $B=2048$ and $R=32$ projection candidates. Each candidate sorts sources and targets along a projection and solves exact quadratic-cost assignments within rank blocks; selection minimizes the full pixel-space coupling cost, including an independent fallback. ReCouple-AS adds annealed spatial regularization with initial penalty-to-CFM loss ratio $\eta=0.25$. ReCouple+ uses one pooled capacitated transport plan with $M=10$ Gaussian sources per target, and ReCouple++ adds the same regularization with $\eta=0.5$. Perturbation magnitude is $0.05$ times the RMS per-coordinate data standard deviation. The penalty coefficient is calibrated on 128 pairs and cosine-annealed to zero over 35,000 updates, followed by 15,000 pure-CFM updates. We also include global exact OT ($B=N$) and Noise-ReCouple, which uses $B=1024$, $R=8$, and replaces each paired source by $\sqrt{0.99}z+\sqrt{0.01}\epsilon$.

\paragraph{Baselines}
We compare against the unchanged pretrained backbone, fresh-independent fine-tuning, exact minibatch OT fine-tuning (OT-CFM-FT) \citep{tong2023improving}, AlignFlow \citep{kong2026alignflow}, and SD-FM \citep{mousavi2026flow}. 
AlignFlow and SD-FM use semidiscrete OT; AlignFlow includes horizontally flipped targets, whereas SD-FM uses the same 50,000 unflipped targets as ReCouple. Fresh-independent and OT-CFM fine-tuning use random horizontal flips. Teacher reflow and one-step distillation share 500,000 Gaussian--endpoint pairs generated by accurately integrating the frozen pretrained teacher, using uniform-time CFM's MSE and endpoint MSE at $t=0$, respectively.

\paragraph{Post-training and evaluation}
Each method receives 50,000 updates with batch size 128, a fresh Adam optimizer, learning rate $2\times10^{-4}$, 5,000-step warmup, gradient clipping at 1, and EMA decay $0.9999$. Training seeds are $\{10,20,30\}$, with fixed pretrained backbones and shared prepared couplings. We evaluate EMA models using uniform Euler integration at NFE $\{1,2,4,8,10,50\}$ and compute FID from 50,000 generated images against CIFAR-10 training statistics using CleanFID's \texttt{legacy\_tensorflow} mode. Fine-tuned models share evaluation seed 9901; results report mean and sample standard deviation across training seeds (see Table \ref{tab:cifar-current-baselines-std}-\ref{tab:cifar-otcfm-baselines-std}). Backbone repetitions instead vary the sampling seed. Distillation results beyond NFE 1 are diagnostic. Preparation and post-training times are reported separately, excluding pretraining and FID evaluation.

\paragraph{Wall-clock cost}

Table~\ref{tab:cifar-wallclock} reports wall-clock costs for coupling preparation and post-training on an NVIDIA GH200. Ordinary ReCouple requires $30.53$ minutes of one-time coupling construction, compared with $542.46$ minutes for the global $B=N$ assignment, a roughly $17.8\times$ reduction in preparation time in our implementation. Its post-training cost remains comparable to the other fine-tuning baselines, at approximately $105$ minutes for either pretrained backbone. Increasing the source resolution through ReCouple+ primarily increases the one-time coupling cost: the extreme $M=10$ pooled construction requires $477.49$ minutes, while its post-training time remains close to that of ordinary ReCouple. In contrast, spatial regularization leaves the coupling unchanged but increases optimization time, as ReCouple-AS and ReCouple++ evaluate the additional consistency objective during post-training. These timings are implementation- and hardware-dependent and are intended to characterize the practical cost of the methods rather than their asymptotic complexity.

\begin{table}[!h]
\centering
\caption{Wall-clock cost comparison for the CIFAR-10 experiments.
Times are reported in minutes. Preparation is the one-time
method-specific cost for constructing the required coupling, pair bank,
or teacher targets and is shared across compatible runs. Post-training
time is reported as mean $\pm$ sample standard deviation over three
training seeds. FID evaluation, backbone pretraining, and scheduler
waiting are excluded. 
${}^\dagger$Teacher preparation uses the recorded
historical allocation wall time with four concurrent GPUs and is
therefore not directly hardware-normalized against the other preparation
times.}
\label{tab:cifar-wallclock}
\vspace{0.1cm}
\small
\setlength{\tabcolsep}{4.0pt}
\begin{tabular}{lrrrr}
\toprule
& \multicolumn{2}{c}{RF backbone}
& \multicolumn{2}{c}{OT-CFM backbone} \\
\cmidrule(lr){2-3}\cmidrule(lr){4-5}
Method
& Prep. & Post-train
& Prep. & Post-train \\
\midrule
Teacher Reflow (MSE)
& $445.53^\dagger$ & $105.89\pm0.61$
& $284.98^\dagger$ & $108.22\pm3.15$ \\
One-step distillation (MSE)
& $445.53^\dagger$ & $107.41\pm3.63$
& $284.98^\dagger$ & $105.41\pm2.73$ \\
\midrule
Fresh-independent FT
& 0.00 & $96.64\pm3.30$
& 0.00 & $96.10\pm3.27$ \\
OT-CFM-FT (\cite{tong2023improving})
& 0.00 & $97.73\pm2.83$
& 0.00 & $97.88\pm1.86$ \\
AlignFlow (\cite{kong2026alignflow})
& 30.41 & $118.53\pm6.19$
& 30.41 & $115.91\pm4.02$ \\
SD-FM (\cite{mousavi2026flow})
& 38.16 & $111.40\pm1.19$
& 38.16 & $111.62\pm3.12$ \\
Noise-ReCouple (cf.\ \citet{hui2025not})
& 2.69 & $108.18\pm2.28$
& 2.69 & $107.17\pm3.41$ \\
ReCouple ($B=N$)
& 542.46 & $96.57\pm2.77$
& 542.46 & $98.52\pm2.90$ \\
ReCouple
& 30.53 & $105.56\pm1.09$
& 30.53 & $104.68\pm3.01$ \\
ReCouple-AS
& 30.53 & $157.12\pm3.19$
& 30.53 & $155.86\pm3.70$ \\
ReCouple+ $(M=10)$
& 477.49 & $105.90\pm3.25$
& 477.49 & $105.95\pm2.99$ \\
ReCouple++ ($M=10)$
& 477.49 & $171.85\pm2.58$
& 477.49 & $165.69\pm1.46$ \\
\bottomrule
\end{tabular}
\end{table}

\begin{table}[!t] 
\centering 
\caption{Sample standard deviation of CIFAR-10 FID with the Rectified Flow pretrained backbone, corresponding to Table~\ref{tab:cifar-current-baselines}. Standard deviations are computed over three seeds. For the RF backbone row, the standard deviation reflects sampling-seed variability; for the fine-tuned methods, it reflects variability across training seeds.} 
\label{tab:cifar-current-baselines-std} \vspace{0.1cm} 
\small \setlength{\tabcolsep}{3.5pt} 
\begin{tabular}{lrrrrrr} \toprule Method & NFE 1 & NFE 2 & NFE 4 & NFE 8 & NFE 10 & NFE 50 \\ \midrule RF (backbone) & 0.166 & 0.291 & 0.174 & 0.115 & 0.094 & 0.011 \\ \midrule Teacher Reflow (MSE) & 0.010 & 0.021 & 0.026 & 0.022 & 0.023 & 0.021 \\ One-step distillation (MSE) & 0.006 & 0.246 & 0.359 & 0.362 & 0.360 & 0.393 \\ \midrule 
Fresh-independent FT & 2.154 & 1.828 & 0.247 & 0.136 & 0.091 & 0.040 \\ OT-CFM-FT (\cite{tong2023improving}) & 0.619 & 0.370 & 0.053 & 0.064 & 0.038 & 0.030 \\ AlignFlow (\cite{kong2026alignflow}) & 0.832 & 0.194 & 0.101 & 0.066 & 0.061 & 0.041 \\ SD-FM (\cite{mousavi2026flow}) & 0.202 & 0.251 & 0.087 & 0.034 & 0.029 & 0.034 \\ Noise-ReCouple (cf.\ \citet{hui2025not}) & 0.226 & 0.099 & 0.121 & 0.135 & 0.121 & 0.147 \\ ReCouple ($B=N$) & 0.634 & 0.288 & 0.150 & 0.049 & 0.033 & 0.051 \\ ReCouple & 0.283 & 0.188 & 0.143 & 0.175 & 0.173 & 0.126 \\ ReCouple-AS & 0.608 & 0.202 & 0.121 & 0.126 & 0.153 & 0.067 \\ ReCouple+ & 0.106 & 0.291 & 0.160 & 0.148 & 0.136 & 0.071 \\ ReCouple++ & 0.748 & 0.008 & 0.081 & 0.124 & 0.113 & 0.081 \\ \bottomrule 
\end{tabular}  
\end{table} 

\begin{table}[!h] \centering \caption{Sample standard deviation of CIFAR-10 FID with the OT-CFM pretrained backbone, corresponding to Table~\ref{tab:cifar-otcfm-baselines}. Standard deviations are computed over three seeds. For the OT-CFM backbone row, the standard deviation reflects sampling-seed variability; for the fine-tuned methods, it reflects variability across training seeds.} 
\label{tab:cifar-otcfm-baselines-std} \vspace{0.1cm} 
\small \setlength{\tabcolsep}{3.5pt} 
\begin{tabular}{lrrrrrr} \toprule Method & NFE 1 & NFE 2 & NFE 4 & NFE 8 & NFE 10 & NFE 50 \\ \midrule OT-CFM (backbone) & 0.141 & 0.035 & 0.161 & 0.040 & 0.023 & 0.010 \\ \midrule Teacher Reflow (MSE) & 0.149 & 0.102 & 0.101 & 0.114 & 0.115 & 0.113 \\ One-step distillation (MSE) & 0.012 & 0.133 & 0.130 & 0.160 & 0.158 & 0.233 \\ \midrule 
Fresh-independent FT & 1.864 & 1.150 & 0.309 & 0.124 & 0.090 & 0.010 \\ OT-CFM-FT (\cite{tong2023improving}) & 0.596 & 0.287 & 0.142 & 0.110 & 0.121 & 0.059 \\ AlignFlow (\cite{kong2026alignflow}) & 1.078 & 0.462 & 0.172 & 0.075 & 0.054 & 0.052 \\ SD-FM (\cite{mousavi2026flow}) & 0.587 & 0.356 & 0.099 & 0.045 & 0.045 & 0.009 \\ Noise-ReCouple (cf.\ \citet{hui2025not}) & 0.719 & 0.066 & 0.061 & 0.153 & 0.154 & 0.142 \\ ReCouple ($B=N$) & 0.278 & 0.216 & 0.151 & 0.101 & 0.075 & 0.005 \\ ReCouple & 0.156 & 0.061 & 0.104 & 0.129 & 0.136 & 0.084 \\ ReCouple-AS & 0.574 & 0.152 & 0.170 & 0.167 & 0.164 & 0.150 \\ ReCouple+ & 0.669 & 0.252 & 0.095 & 0.060 & 0.030 & 0.016 \\ ReCouple++ & 0.612 & 0.340 & 0.105 & 0.038 & 0.048 & 0.046 \\ \bottomrule 
\end{tabular}
\end{table}

\subsection{Ablation Studies}

Tables~\ref{tab:cifar-hparam-ablation-low}--\ref{tab:cifar-hparam-ablation-high}
report generation quality across numerical resolutions, while
Table~\ref{tab:cifar-coupling-ablation} summarizes the corresponding
coupling-side diagnostics.

\paragraph{Block size and projection search}
Increasing $B$ systematically improves the endpoint transport objective:
the normalized gap to global empirical OT decreases from $0.256$ at
$B=512$ to $0.211$ at $B=1024$ and $0.169$ at $B=2048$.  The
downstream benefit is concentrated at very low NFEs: NFE-1 FID decreases
from $137.42\pm1.04$ to $130.07\pm0.26$ and $123.16\pm0.28$, and
NFE-2 FID from $48.89\pm0.36$ to $47.02\pm0.05$ and
$45.72\pm0.19$.  From NFE~4 onward, however, $B=1024$ and $B=2048$
perform similarly.  Thus, reducing the endpoint-cost gap to empirical OT
does not translate uniformly into improved generation at every numerical
resolution.

Increasing $R$ from $4$ to $16$ or $32$ has no effect for any tested
$B$: the assignment hashes are identical, and hence so are the coupling
diagnostics and downstream FIDs.  Since the proposal sets are nested,
the selected candidate is already present among the first four
projections.  Thus best-of-$R$ search has saturated at $R=4$ for these
endpoint banks, although this need not hold in general.

\paragraph{Source resolution}
The pooling factor $M$ produces a pronounced NFE-dependent crossover.
At fixed $B=1024$ and $R=16$, the unregularized family is best at
$M=1$ for NFE~1--2, $M=2$ for NFE~4, $M=5$ for NFE~8, and $M=10$
for NFE~10--50.  For example, increasing $M$ from $2$ to $10$ worsens
NFE-1 FID from $155.74\pm0.32$ to $195.09\pm0.73$, while improving
NFE-50 FID from $8.28\pm0.07$ to $5.23\pm0.01$.

By contrast, the coupling-side objective changes only modestly:
relative cost reduction increases from $3.91\%$ at $M=2$ to $3.96\%$
at $M=5$ and $3.97\%$ at $M=10$.  Hence endpoint quadratic cost alone
does not explain the large change in the FID profile.  The results are
instead consistent with the role of $M$ as a source-resolution
parameter: finer source resolution can improve generation at
larger NFEs while being detrimental in the few-step regime.

\paragraph{Spatial regularization}
Here, $\eta$ specifies the initial regularization strength relative
to the CFM loss: we calibrate $\lambda_0$ on a fixed batch so that
$\lambda_0 \mathcal{L}_{\mathrm{AS}} /
\mathcal{L}_{\mathrm{CFM}} = \eta$ at initialization.
Thus, $\eta$ is a loss ratio, rather than the regularization
coefficient itself; $\eta=0$ disables regularization.

The $\eta$ sweep cleanly separates neural realization from coupling
construction.  All ReCouple++ settings use exactly the same $M=5$
pooled plan, so the coupling diagnostics are unchanged as $\eta$
varies.  Nevertheless, increasing $\eta$ monotonically improves mean
FID from NFE~4 onward.  Relative to unregularized ReCouple+ at $M=5$,
$\eta=1$ reduces FID from $23.25\pm0.16$ to $22.75\pm0.15$ at
NFE~4, from $12.61\pm0.11$ to $11.94\pm0.17$ at NFE~8, from
$11.04\pm0.11$ to $10.33\pm0.15$ at NFE~10, and from
$5.76\pm0.04$ to $5.00\pm0.13$ at NFE~50.  Its effect at NFE~1--2
is comparatively small.  Since the endpoint plan is fixed, these gains
arise downstream of coupling construction, through how the transport
field is learned and represented.

\begin{table}[!t]
\centering
\caption{
CIFAR-10 ablations with the RF backbone at low NFEs.
For ReCouple, we jointly vary the block size $B$ and number of projection
proposals $R$. For ReCouple+ and ReCouple++, we vary the pooling factor $M$
and spatial-regularization strength $\eta$, respectively, while fixing the
remaining hyperparameters to their default values.
Each cell reports FID as mean $\pm$ sample standard deviation over three seeds
(lower is better).
}
\label{tab:cifar-hparam-ablation-low}
\vspace{0.1cm}
\small
\setlength{\tabcolsep}{4.0pt}
\begin{tabular}{llrrrrrrr}
\toprule
Method & Ablation
& $B$ & $M$ & $R$ & $\eta$
& NFE 1 & NFE 2 & NFE 4 \\
\midrule

\multirow{9}{*}{ReCouple}
& \multirow{9}{*}{$(B,R)$}
& 512  & 1 & 4  & 0
& $137.42\pm1.04$
& $48.89\pm0.36$
& $23.48\pm0.13$ \\
&& 512  & 1 & 16 & 0
& $137.42\pm1.04$
& $48.89\pm0.36$
& $23.48\pm0.13$ \\
&& 512  & 1 & 32 & 0
& $137.42\pm1.04$
& $48.89\pm0.36$
& $23.48\pm0.13$ \\
\cmidrule(lr){3-9}
&& 1024 & 1 & 4  & 0
& $130.07\pm0.26$
& $47.02\pm0.05$
& $22.99\pm0.12$ \\
&& 1024 & 1 & 16 & 0
& $130.07\pm0.26$
& $47.02\pm0.05$
& $22.99\pm0.12$ \\
&& 1024 & 1 & 32 & 0
& $130.07\pm0.26$
& $47.02\pm0.05$
& $22.99\pm0.12$ \\
\cmidrule(lr){3-9}
&& 2048 & 1 & 4  & 0
& $123.16\pm0.28$
& $45.72\pm0.19$
& $23.01\pm0.14$ \\
&& 2048 & 1 & 16 & 0
& $123.16\pm0.28$
& $45.72\pm0.19$
& $23.01\pm0.14$ \\
&& 2048 & 1 & 32 & 0
& $123.16\pm0.28$
& $45.72\pm0.19$
& $23.01\pm0.14$ \\

\midrule

\multirow{3}{*}{ReCouple+}
& \multirow{3}{*}{$M$}
& 1024 & 2  & 16 & 0
& $155.74\pm0.32$
& $52.72\pm0.05$
& $21.38\pm0.03$ \\
&& 1024 & 5  & 16 & 0
& $188.44\pm0.39$
& $66.81\pm0.42$
& $23.25\pm0.16$ \\
&& 1024 & 10 & 16 & 0
& $195.09\pm0.73$
& $71.41\pm0.07$
& $24.59\pm0.22$ \\

\midrule

\multirow{3}{*}{ReCouple++}
& \multirow{3}{*}{$\eta$}
& 1024 & 5 & 16 & 0.25
& $188.16\pm0.58$
& $66.59\pm0.53$
& $23.05\pm0.12$ \\
&& 1024 & 5 & 16 & 0.50
& $187.91\pm0.55$
& $66.68\pm0.17$
& $22.95\pm0.16$ \\
&& 1024 & 5 & 16 & 1.00
& $187.94\pm0.55$
& $66.56\pm0.55$
& $22.75\pm0.15$ \\

\bottomrule
\end{tabular}
\end{table}

\begin{table}[!t]
\centering
\caption{
CIFAR-10 ablations with the RF backbone at higher NFEs.
Experimental settings are identical to
Table~\ref{tab:cifar-hparam-ablation-low}.
Each cell reports FID as mean $\pm$ sample standard deviation over three seeds
(lower is better).
}
\label{tab:cifar-hparam-ablation-high}
\vspace{0.1cm}
\small
\setlength{\tabcolsep}{4.0pt}
\begin{tabular}{llrrrrrrr}
\toprule
Method & Ablation
& $B$ & $M$ & $R$ & $\eta$
& NFE 8 & NFE 10 & NFE 50 \\
\midrule

\multirow{9}{*}{ReCouple}
& \multirow{9}{*}{$(B,R)$}
& 512  & 1 & 4  & 0
& $16.99\pm0.17$
& $15.94\pm0.20$
& $11.49\pm0.16$ \\
&& 512  & 1 & 16 & 0
& $16.99\pm0.17$
& $15.94\pm0.20$
& $11.49\pm0.16$ \\
&& 512  & 1 & 32 & 0
& $16.99\pm0.17$
& $15.94\pm0.20$
& $11.49\pm0.16$ \\
\cmidrule(lr){3-9}
&& 1024 & 1 & 4  & 0
& $16.53\pm0.08$
& $15.45\pm0.08$
& $10.95\pm0.11$ \\
&& 1024 & 1 & 16 & 0
& $16.53\pm0.08$
& $15.45\pm0.08$
& $10.95\pm0.11$ \\
&& 1024 & 1 & 32 & 0
& $16.53\pm0.08$
& $15.45\pm0.08$
& $10.95\pm0.11$ \\
\cmidrule(lr){3-9}
&& 2048 & 1 & 4  & 0
& $16.62\pm0.18$
& $15.54\pm0.17$
& $10.95\pm0.13$ \\
&& 2048 & 1 & 16 & 0
& $16.62\pm0.18$
& $15.54\pm0.17$
& $10.95\pm0.13$ \\
&& 2048 & 1 & 32 & 0
& $16.62\pm0.18$
& $15.54\pm0.17$
& $10.95\pm0.13$ \\

\midrule

\multirow{3}{*}{ReCouple+}
& \multirow{3}{*}{$M$}
& 1024 & 2  & 16 & 0
& $13.84\pm0.01$
& $12.68\pm0.03$
& $8.28\pm0.07$ \\
&& 1024 & 5  & 16 & 0
& $12.61\pm0.11$
& $11.04\pm0.11$
& $5.76\pm0.04$ \\
&& 1024 & 10 & 16 & 0
& $12.67\pm0.02$
& $10.93\pm0.01$
& $5.23\pm0.01$ \\

\midrule

\multirow{3}{*}{ReCouple++}
& \multirow{3}{*}{$\eta$}
& 1024 & 5 & 16 & 0.25
& $12.40\pm0.09$
& $10.85\pm0.11$
& $5.49\pm0.07$ \\
&& 1024 & 5 & 16 & 0.50
& $12.28\pm0.09$
& $10.68\pm0.10$
& $5.31\pm0.08$ \\
&& 1024 & 5 & 16 & 1.00
& $11.94\pm0.17$
& $10.33\pm0.15$
& $5.00\pm0.13$ \\

\bottomrule
\end{tabular}
\end{table}

\begin{table}[!h]
\centering
\caption{
Coupling diagnostics for the CIFAR-10 RF ablations.
Relative cost reduction is
$100(1-C/C_{\mathrm{ind}})$, where $C$ is the average sum of squared
pixelwise endpoint differences and $C_{\mathrm{ind}}$ is the cost of
independent pairing using the same source pool and empirical target
marginal for each $M$.
For $M=1$, we additionally report the normalized endpoint-cost gap to
global empirical OT,
$(C-C_{\mathrm{OT}})/(C_{\mathrm{ind}}-C_{\mathrm{OT}})$.
For each $B$, $R\in\{4,16,32\}$ yields the same assignment; all
ReCouple++ settings share the $M=5$ ReCouple+ plan.
}
\label{tab:cifar-coupling-ablation}
\vspace{0.1cm}
\small
\setlength{\tabcolsep}{8pt}
\begin{tabular}{llrr}
\toprule
Method & Setting
& Rel.\ cost reduction (\%) $\uparrow$
& Norm.\ empirical-OT gap $\downarrow$ \\
\midrule
ReCouple  & $B=512$  & 3.62 & 0.256 \\
ReCouple  & $B=1024$ & 3.84 & 0.211 \\
ReCouple  & $B=2048$ & 4.04 & 0.169 \\
\midrule
ReCouple+ & $M=2$  & 3.91 & -- \\
ReCouple+ & $M=5$  & 3.96 & -- \\
ReCouple+ & $M=10$ & 3.97 & -- \\
\bottomrule
\end{tabular}
\end{table}

{\bf Takeaway.} Taken together, the ablations separate four roles as follows.
Increasing $B$ reduces the endpoint-cost gap to empirical OT and is most
beneficial in the few-step regime; increasing $M$ instead induces a
clear coarse-to-fine trade-off across NFEs; and spatial regularization
can improve a fixed pooled coupling without altering its endpoint cost.
Overall, the results show that endpoint transport cost alone is
insufficient to characterize a useful coupling for flow matching:
downstream quality depends on how coupling resolution, neural
realization, and numerical resolution interact.

\subsection{ImageNet-256 Implementation Details}
\label{app:imagenet_impl}

All experiments use the class-conditional SiT-XL/2 model initialized from the same pretrained checkpoint and operate on ImageNet-256 MSE-VAE latents. Post-training uses AdamW with learning rate $10^{-4}$, four warmup steps, cosine decay with minimum learning-rate ratio $0.1$, gradient clipping at $1.0$, and EMA decay $0.999$. We use a  batch size of $8$ and eight gradient-accumulation steps, yielding an effective batch size of $64$. Each coupling plan contains $6{,}000$ updates. Based on NFE-$5$ validation FID, we select the $2$K ReCouple checkpoint, $3$K ReCouple-AS and ReCouple++ checkpoints, $6$K minibatch-OT checkpoint, and $3$K SD-FM checkpoint.

ReCouple uses class-conditional coupling pools of size $4{,}096$. ReCouple++ uses pooling factor $M=2$. Minibatch OT computes assignments within each batch of $64$. For SD-FM, we solve one unregularized semi-discrete OT problem per ImageNet class using assignments
\[
j^\star(z)=\arg\max_j\{\langle z,y_j\rangle+g_j\}.
\]
The potentials are optimized with AdaGrad for $1{,}500$ steps using source batches of $256$, with a constant schedule for the first $1{,}000$ steps and averaging over the final $250$ steps. Assignment balance is estimated using $65{,}536$ source samples, after which we construct a fixed $6{,}000\times64$ pairing plan using assignment batches of $128$.

For evaluation, we generate $50{,}000$ samples using Euler integration, classifier-free guidance scale $4.0$, per-process batch size $8$, four GPUs, and global seed $123$. An evaluation at NFE $K$ uses $K+1$ time-grid points and $K$ Euler updates. All metrics are computed against the same ImageNet-256 reference statistics using the guided-diffusion evaluator.

\end{document}